\documentclass{elsart}
\usepackage{algorithmic}
\usepackage{algorithm}
\usepackage{graphicx}
\usepackage{amssymb}
\usepackage{amsmath}
\usepackage{enumerate}

\usepackage{multirow} 
\usepackage{booktabs}
\usepackage[switch,pagewise]{lineno}
\usepackage[usenames]{color}
\usepackage{url}
\usepackage{caption}
\usepackage{lineno}

\usepackage{verbatim}
\usepackage{float}
\usepackage{subfigure}
\usepackage{adjustbox}
\usepackage{color}
\usepackage{cite}
\usepackage{appendix}
\usepackage{array}
\usepackage{caption}
\usepackage{mathrsfs} 

\usepackage{threeparttable}
\usepackage{supertabular}
\usepackage{longtable}
\usepackage{lscape}
\usepackage{bbding}

\graphicspath{{EPS/}{PDF/}}
\newcommand{\doi}[1]{\textsc{doi}: \href{http://dx.doi.org/#1}{\nolinkurl{#1}}}

\newtheorem{mypro}{Proposition}
\begin{document}

\begin{frontmatter}

\title{An effective hybrid search algorithm for the multiple traveling repairman problem with profits}

\author[Shenzhen,Anhui]{Jintong Ren}, 
\ead{renjintong@cuhk.edu.cn}
\author[Angers]{Jin-Kao Hao},
\ead{jin-kao.hao@univ-angers.fr}
\author[Anhui]{Feng Wu}
\ead{wufeng02@ustc.edu.cn} \and
\author[Shenzhen,Airs]{Zhang-Hua Fu\corauthref{cor}}
\corauth[cor]{Corresponding author.}
\ead{fuzhanghua@cuhk.edu.cn}

\address[Shenzhen]{The Chinese University of Hong Kong, Shenzhen 518172, P.R. China}
\address[Anhui]{University of Science and Technology of China, Hefei 230026, P.R.China}
\address[Angers]{LERIA, Universit{\'e} d'Angers, 2 boulevard Lavoisier, 49045 Angers, France}
\address[Airs]{Shenzhen Institute of Artificial Intelligence and Robotics for Society, Shenzhen 518172, P.R. China}
\maketitle

\begin{abstract}
As an extension of the traveling repairman problem with profits, the multiple traveling repairman problem with profits consists of multiple repairmen who visit a subset of all customers to maximize the revenues collected through the visited customers.
To solve this challenging problem, an effective hybrid search algorithm based on the memetic algorithm framework is proposed. 
It integrates two distinguished features: a dedicated arc-based crossover to generate high-quality offspring solutions and a fast evaluation technique to reduce the complexity of exploring the classical neighborhoods.
%
%
%
%
We show the competitiveness of the algorithm on 470 benchmark instances compared to the leading reference algorithms and report new best records for 137 instances as well as equal best results for other 330 instances. 
We investigate the importance of the key search components for the algorithm.

\end{abstract}
\end{frontmatter}


\section{Introduction}
\label{sec:Intro}

The traveling repairman problem with profits (TRPP) \cite{dewilde_heuristics_2013} is a general model that can be stated as follows. Let $G(V,E)$ be a complete weighted graph, where $V$ is the vertex set consisting of the depot $0$ and the customer set $V_c=\{1, 2, ..., n\}$,  and $E=\{(i,j): i,j \in V \}$ is the edge set where each edge $(i,j)$ is associated with a symmetric weight $d_{i,j}=d_{j,i}$ (traveling time). A repairman begins his trip from the depot to collect a time-dependent revenue $p_i-l(i)$ via each visited customer and stops his travels when there is no positive revenue, where $p_i$ represents the profit and $l(i)$ is the waiting time for each customer $i$ ($l(0)=0$). Each customer can be visited at most once. The objective of the TRPP is to find an open Hamiltonian path such that the collected revenue
$\sum_{i=0}^m [p_i-l(i)]^+$ is maximized,
 where $m$ is the number of visited customers  and $[p_i-l(i)]^+$ is the larger value between $p_i-l(i)$ and $0$.

The multiple traveling repairman problem with profits (MTRPP) generalizes the TRPP by considering multiple repairmen (servers or vehicles) to serve the customers. In the MTRPP, all the repairmen start their trips from the depot to collect a time-dependent revenue independently. Let $K\geq 1$ be the number of repairmen, then a formal solution $\varphi$ consists of $K$ Hamiltonian paths (or routes) $\{X_1, X_2, ..., X_{K}\}$,  where each path $X_k=(x_0^k, x_1^k, ..., x_{m_k}^k)$ contains $m_k$ customers ($\bigcup\limits_{k=1}^{K} X_i \subseteq V$ and $X_i \cap X_j$=$\{0\},$ $i\neq j$,  $\forall i,j \in \{1, 2, ..., K\}$). The objective function can be defined as follows.
\begin{equation}\label{eq:definition}
f(\varphi)=\sum_{k=1}^{K} \sum_{i=0}^{m_k} [p_{x_i^k}-l(x_i^k)]^+
\end{equation}

The aim of the MTRPP is then to find the solution $\varphi^*$ with a maximal total collected revenue $f(\varphi^*)$.

If none of the collected revenues $p_{x_i^k}-l(x_i^k)$ is negative, then Equation~(\ref{eq:definition}) can be rewritten as follows.
\begin{equation}\label{eq:fast}
f(\varphi)=\sum_{k=1}^{K} \sum_{i=0}^{m_k} p_{x_i^k}- \sum_{k=1}^{K} \sum_{i=1}^{m_k} (m_k-i+1)\cdot d_{x^k_{i-1}, x^k_{i}}
\end{equation}
Equation~(\ref{eq:fast}) is useful to design the fast evaluation of our search algorithm. 

Typical applications of the MTRPP in the real-life concern humanitarian and emergency relief logistics. For instance, for the case of post-disaster relief operations, $K$ homogeneous rescue teams start their trips from the base to deliver emergency supplies and save survivors of each damaged village or city. Assuming that $p_i$ persons need to be rescued for the village $i$ and one person loses his life along with each time step. The objective of the rescue teams is to save as many as lives. This application scenario was also mentioned for the TRPP \cite{dewilde_heuristics_2013} with a single rescue team. Clearly, the MTRPP is a more convenient model for the real situation where several rescue teams are needed.

Existing work for solving the MTRPP as well as some related problems are briefly reviewed as follows.

As to solution methods for the MTRPP, there are two practical algorithms in the literature.
In 2019, Lu et al. \cite{lu2019memetic} proposed the first memetic algorithm (MA-MTRPP) to solve the MTRPP. This algorithm uses a randomized greedy construction phase for solution initialization, a variable neighborhood search for local optimization and a route-based crossover operator for solution recombination. MA-MTRPP showed to be more efficient compared to the general CPLEX solver on the 240 benchmark instances introduced in that paper.
%
%
In the same year, Avci and Avci \cite{avci_adaptive_2019} developed a mixed-integer linear programming model and suggested an adaptive large neighborhood algorithm search approach (ALNS-MTRPP) for the MTRPP, which incorporates a couple of problem-specific destroy operators and two new randomized repair operators. The authors proposed another set of 230 benchmark instances and a greedy randomized adaptive search procedure with iterated local search (GRASP-ILS), which was used as a reference heuristic. According to the experimental results, ALNS-MTRPP appears to be more effective and time-efficient than GRASP-ILS for most instances.

The closely related TRPP is a special case of the MTRPP with a single repairman ($K=1$). Several heuristic algorithms were proposed to solve the TRPP. In 2013, Dewilde et al. \cite{dewilde_heuristics_2013} firstly proposed a tabu search algorithm incorporating multiple neighborhoods and a greedy initialization procedure. In 2017, Avci and Avci \cite{avci_grasp_2017} suggested a greedy randomized adaptive search procedure combined with  iterated local search, which outperformed the previous algorithms by updating 46 best results. In 2019, Lu et al. \cite{lu_hybrid_2019} introduced a population-based hybrid evolutionary search algorithm which provided better results than the previous algorithms. In 2020, Pei et al. \cite{pei_solving_2020} developed a general variable neighborhood search approach integrating auxiliary data structures to improve the search efficiency. This algorithm dominated all the previous algorithms by updating 40 best-known results and matching the best-known results for the remaining instances. As we show in the current work, these auxiliary data structures can be beneficially extended to the MTRPP to design fast evaluation techniques for the generalized problem. 

The team orienteering problem (TOP) \cite{chao1996team} is another related problem that states that a fixed number of homogeneous vehicles visit a subset of customers to maximize the collected profits within a traveling distance constraint. Different from the MTRPP, the profits of customers in the TOP are time-independent and there are distance constraints for the vehicles. Various solution methods were developped for the TOP, including local search algorithms \cite{vansteenwegen2009guided, hammami2020hybrid, tsakirakis2019similarity},  population based algorithms \cite{bouly2010memetic, zettam2016novel, dang2013effective} and exact methods based on branch-and-price and the cutting plane technique \cite{boussier2007exact, bianchessi2018branch, el2016solving, poggi2010team}. The cumulative capacitated vehicle routing problem (CCVRP) is also related to the MTRPP by considering capacity constraints for the $K$ repairmen (or vehicles). Popular algorithms for this problem include evolutionary algorithm \cite{ngueveu2010effective}, adaptive large neighborhood search heuristic \cite{ribeiro2012adaptive}, two-phase metaheuristic \cite{ke2013two}, iterated greedy algorithms \cite{nucamendi2018cumulative}, and brand-and-cut-and-price algorithm \cite{lysgaard2014branch}.

The MTRPP (with multiple repairmen) is a more realistic model compared to the TRPP (with a single repairman) for real-life applications. However, there are only two principal heuristics designed for the MTRPP, contrary to the case of the TRPP for which numerous solution methods exist. There is thus a need to enrich the solution tools for this relevant problem. Moreover, the two existing algorithms for the MTRPP are rather sophisticated and rely on many parameters (13 parameters for ALNS-MTRPP and 7 for MA-MTRPP). In addition, ALNS-MTRPP has difficulties in handling instances of large-scale (e.g., requiring 3 hours to solve the 1000-customers instances). 

In this work, we propose an easy-to-use (with only 3 parameters) and effective hybrid search algorithm based on the memetic framework to solve the MTRPP (named EHSA-MTRPP). We summarize the contributions as follows. 

First, we propose an original arc-based crossover (\emph{ABX}), which is inspired by  experimental observation and backbone-based heuristics \cite{zhang2004configuration, wang2013backbone}. ABX is able to generate promising offspring solutions from high quality parent solutions.

Second, to ensure a high computational effectiveness, we introduce a series of data structures to reduce the complexities for the examination of the neighborhoods and prove that evaluating one neighboring solution in the underlying neighborhoods for the MTRPP can be performed in constant time.
%
%

Third, we provide new lower bounds for 137 instances out of the 470 benchmark instances in the literature. These bounds can be used for future studies on the MTRPP. 

Finally, we will make the source code of the proposed algorithm publicly available to the community. People working on the MTRPP and related applications can use the code to solve their problems. This fills the gap that no code for solving the MTRPP is currently available.


In the next section, we present the proposed algorithm. In Section~\ref{sec:Experi}, we show the experimental setup, parameter tuning and computational results, followed by investigation of the key component of  the algorithm in Section~\ref{sec:Analyse}. We draw conclusions and provides research perspectives in the last Section~\ref{sec:Conclu}.

\section{Method}
\label{sec:Algo}

\subsection{Main scheme}\label{subsec:Algo:main}

The proposed hybrid search algorithm for the MTRPP is based on the framework of the memetic algorithm \cite{moscato1999memetic} and relies on five search components: a population initialization procedure (\emph{IniPool}), a variable neighborhood search procedure (\emph{VNS}) to perform the local refinement, a perturbation procedure (\emph{Spert}) to help escape from the local optimum, an arc-based crossover (\emph{ABX}) to generate high-quality offspring solutions and a pool updating procedure (\emph{UpdatingPool}) to manage the population with newly obtained solutions.


\begin{algorithm}[!tb]
	\begin{small}
		\caption{The general scheme of the EHSA-MTRPP algorithm}
		\label{Algo:EHSA-MTRPP}
		\begin{algorithmic}[1]
			\renewcommand{\algorithmiccomment}[1]{\hfill\textrm{// \small{#1}}}
			\STATE \sf \textbf{Input}: Input graph $G(V,E)$, population size $Nump$, search limit $Limi$, objective function $f$ and maximum allowed time $T_{max}$
			\STATE \textbf{Output}: Best found solution $\varphi^*$
			
			\STATE /* \emph{IniPool} is used to generate initial population. */
			\STATE /* \emph{VNS} is used to perform the local refinement. */		
			\STATE /* \emph{Spert} is used to modify (slightly) the input local optimum. */		
			\STATE /* \emph{ABX} is used to generate promising offspring solutions. */
			\STATE /* \emph{UpdatingPool} is used to update the population. */				
				
			\STATE $P=\{\varphi_1, ... \varphi_p\} \leftarrow$ \emph{IniPool}()
			\COMMENT{See Section \ref{subsec:Algo:Greedy}}
			
			\FOR{$i \gets 1$ to $Nump$}
			\STATE $\varphi_i \leftarrow$ \emph{VNS}($\varphi_i$)
			\COMMENT{See Section \ref{subsec:Algo:VNS}}
			\ENDFOR
			\STATE $\varphi^*\leftarrow \arg\max\{f(\varphi_i), i=1, ..., Nump\} $
						
			\WHILE{$T_{max}$ is not reached}
			\STATE $C\leftarrow 0$
			\STATE ($\varphi_a, \varphi_b$)$\leftarrow$ RandomChoose($P$)
			
			\STATE $\varphi\leftarrow$ \emph{ABX($\varphi_a, \varphi_b$)}
			\COMMENT{See Section \ref{subsec:Algo:abx}}
			\STATE $\varphi_{lb}\leftarrow \varphi$
			\REPEAT 
			\STATE $\varphi \leftarrow$ \emph{VNS}($\varphi$)

			\IF{$f(\varphi)>(\varphi_{lb})$}
			\STATE $\varphi_{lb} \leftarrow \varphi$
			\STATE $C\leftarrow 0$
			\ELSE
			\STATE $C\leftarrow C+1$
			\ENDIF			
			
			\STATE $\varphi\leftarrow$ \emph{Spert}($\varphi$)
			\COMMENT{See Section \ref{subsec:Algo:Sperb}}
			\UNTIL{$C\geq Limi$}			
			
			\STATE \emph{UpdatingPool}($\varphi_{lb}, P$)
			\COMMENT{See Section \ref{subsec:Algo:updating}}			
		
			\IF{$f(\varphi_{lb})>(\varphi^*)$}
			\STATE $\varphi^* \leftarrow \varphi_{lb}$
			\ENDIF

			\ENDWHILE
			\RETURN $ \varphi^*$
		\end{algorithmic}
	\end{small}
\end{algorithm}

Algorithm~\ref{Algo:EHSA-MTRPP} shows the general scheme of the EHSA-MTRPP algorithm. 
At first, the algorithm calls \emph{IniPool} (See Section~\ref{subsec:Algo:Greedy}) to create the population  $P$, where each solution $\varphi_i$ is improved by \emph{VNS} (See Section~\ref{subsec:Algo:VNS}) and the best one is recorded in $\varphi^*$ (lines 8-12). 
Then the algorithm enters the main search procedure (lines 13-32). For the while loop, we set $C$ to 0 (line 14), randomly choose two different solutions $\varphi_a$ and $\varphi_b$ from the population $P$ and generate an offspring solution $\varphi$ (lines 15-16) with \emph{ABX} (See Section~\ref{subsec:Algo:abx}).
%
After recording $\varphi$ by $\varphi_{lb}$, the algorithm enters the inner loop (lines 18-27) to explore the new solutions by iterating the \emph{VNS} procedure and the \emph{Spert} procedure. For each inner loop, the current solution $\varphi$ is first improved by \emph{VNS} (line 19) and then used to update the local best solution $\varphi_{lb}$. If $\varphi$ is better than $\varphi_{lb}$, $\varphi_{lb}$ is updated and the counter $C$ is reset to 0 (lines 20-22). Otherwise, $C$ is incremented by 1 (lines 23-25). Then the perturbation procedure \emph{Spert} is triggered to displace the search from the local optimum (line 28). The above procedures are repeated until $C$ reaches the search limit $Limi$ (line 27), indicating that the search is exhausted (and trapped in a deep local optimal solution). 
After the inner loop, the local best solution $\varphi_{lb}$ is used to upgrade the population (line 28), and to update the best found solution $\varphi^*$ (lines 29-31). 
When the cutoff time $T_{max}$ is reached (line 13), the whole algorithm stops and returns the best recorded solution $\varphi^*$ (line 33).  

\subsection{Initial population}\label{subsec:Algo:Greedy}

The initial population is filled with two types of solutions: half of them are created with a randomized construction method while the remaining solutions are generated with a greedy construction method.

For the randomized construction method, we first create a giant tour with all the customers in a random order. Then we separate the giant tour into $K$ routes, where each route has the same number of customers. This leads to a complete solution $\varphi$.

We also employed the greedy construction method of \cite{avci_adaptive_2019}.  Starting from an empty solution $\varphi$ with $K$ routes and a vertex list $V_r=\{ 1, 2, ..., n\}$, the greedy construction method iteratively adds one vertex into the solution following a greedy randomized principle. At each step, we evaluate the objective variation of the solution $\varphi$ for each operation $Ope(v, k)$,  which represents adding $v\in V_r$ to the route $k$.  Then we construct a candidate set $OPE_c$ consisting of the $q$ operations with the largest contributions to the objective value. At last, a random operation $Ope(v, k)\in OPE_c$ is carried out to extend the partial solution and the vertex $v$ is removed from $V_r$. These steps are repeated until all the customers are added into the solution. The parameter $q$ is set to 3 here. More details, please refer to \cite{avci_adaptive_2019}.

\subsection{Solution improvement by variable neighborhood search}\label{subsec:Algo:VNS}

For local optimization, we adopt the general Variable Neighborhood Search (VNS) method \cite{mladenovic1997variable}. Indeed, this method has proved to be quite successful for both the TRPP \cite{pei_solving_2020, lu_hybrid_2019, avci_grasp_2017} and the MTRPP \cite{lu2019memetic}. Our \emph{VNS} procedure for the MTRPP  is presented in Algorithm~\ref{Algo:VNS}. 

\begin{algorithm}[!tb]
	\begin{small}
		\caption{Local optimization with VNS} \label{Algo:VNS}
		\begin{algorithmic}[1]
			\renewcommand{\algorithmiccomment}[1]{\hfill\textrm{// \small{#1}}}
			\STATE \sf \textbf{Input}: Objective function $f$ and current solution $\varphi$
			\STATE \textbf{Output}: Local best solution $\varphi$
			
			\STATE /* $N_1, N_2, N_3, N_4$ represent respectively $Swap$, $Insert$, $2\mbox{-}opt$ and $Or\mbox{-}opt$ neighborhoods. */
			\STATE /* $N_5, N_6, N_7$ represent respectively $Inter\mbox{-}Swap$, $Inter\mbox{-}Insert$ and $Inter\mbox{-}2\mbox{-}opt$ neighborhoods. */
			\STATE /* $N_{Add}, N_{Drop} $ denote $Add$ and $Drop$ neighborhoods. */
			
			\REPEAT
			\STATE $\varphi^\prime\leftarrow$ $\varphi$
			\STATE $S_N\leftarrow \{N_1,N_2,N_3,N_4,N_5, N_6, N_7\}$
			\STATE $\varphi \leftarrow LocalSearch(\varphi, N_{Add})$		
			\WHILE{$S_N\neq\emptyset$}
			\STATE Randomly choose a neighborhood $N\in S_N$
			\STATE $\varphi\leftarrow LocalSearch(\varphi, N) $
			\STATE $\varphi \leftarrow LocalSearch(\varphi, N_{Drop})$
			\STATE $S_N\leftarrow S_N\setminus \{N\}$
			\ENDWHILE						
			\UNTIL{$f(\varphi^\prime)\geq f(\varphi)$}
			
			\RETURN $ \varphi$
		\end{algorithmic}
	\end{small}
\end{algorithm}

In the outer loop (lines 6-16), we firstly initialize the recorded solution $\varphi^\prime$ with the current solution $\varphi$ and the neighborhood set $S_N$ with 7 different neighborhoods $N_1\mbox{-}N_7$ (lines 7-8). After a local search procedure based on $N_{Add}$ with the current solution (line 9), the search enters the inner loop to explore local best solutions by alternating different neighborhoods (lines 10-15). For each inner loop, we randomly choose a neighborhood $N\in S_N$ and use it to carry out a local optimization from the current solution (lines 11-12). Then, an additional local optimization based on $N_{Drop}$ is performed and the neighborhood $N$ is removed from the neighborhood set $S_N$ (lines 13-14). When the neighborhood set $S_N$ has been explored ($S_N =\emptyset$), the inner loop is ended. These steps are repeated until no improving solution exists in the neighborhoods (line 16). And $\varphi$ is returned (line 17). 


Our \emph{VNS} procedure exploits three sets of 9 neighborhoods where seven of them are also employed in \cite{lu2019memetic, avci_adaptive_2019}. The first set of four neighborhoods changes the order of customers in one route:
\begin{itemize}
	\item[$\bullet$] $Swap$ ($N_1$): Exchanging the visiting positions of two customers in one route. 
	\item[$\bullet$] $Insert$ ($N_2$): Removing one customer from its position and inserting it into two adjacent nodes in the same route. 
	\item[$\bullet$] $2\mbox{-}opt$ ($N_3$): Removing two non-adjacent edges and replacing them with two new edges in the same route. 
	\item[$\bullet$] $Or\mbox{-}opt$ ($N_4$): Removing a block of $h$ ($h=2,3$) consecutive customers from one route and inserting them into two adjacent nodes in the same route. 
\end{itemize}

\begin{figure}
	\centering
	\includegraphics[scale=0.3]{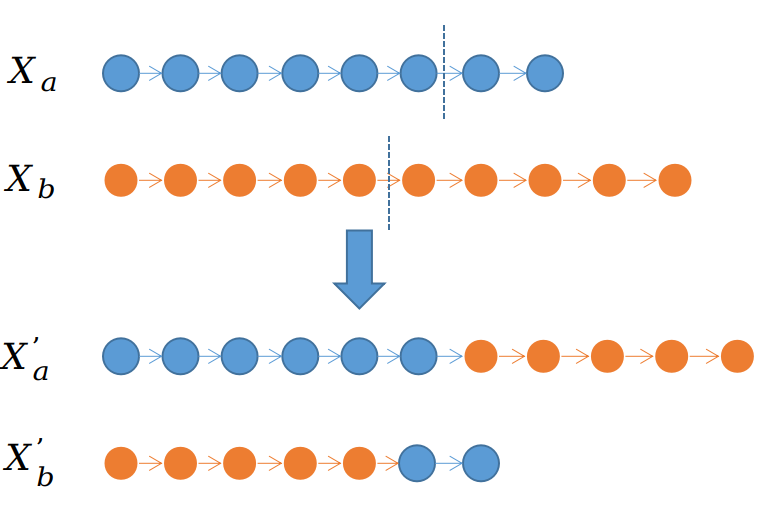}
	\caption{Illustration of $Inter\mbox{-}2\mbox{-}opt$: supposing two routes $X_a$ (marked in blue) and $X_b$ (marked in orange), operating an $Inter\mbox{-}2\mbox{-}opt$ produces two new routes $X_a^\prime$ and $X_b^\prime$, where the blue dotted lines represent the edges to remove.}
	\label{fig:inter2opt}
\end{figure}

The second set of three neighborhoods is designed to change the customers between different routes:
\begin{itemize}
	\item[$\bullet$] $Inter\mbox{-}Swap$ ($N_5$): Exchanging the positions of two customers in two different routes. 
	\item[$\bullet$] $Inter\mbox{-}Insert$ ($N_6$): Removing one customer from one route and inserting it into two adjacent nodes in another route. 
	\item[$\bullet$] $Inter\mbox{-}2\mbox{-}opt$ ($N_7$): Removing two edges from two different routes and replacing them with two new edges. A simple illustration is presented in Figure~\ref{fig:inter2opt}.
\end{itemize}

The third set of two neighborhoods is to change the set of visited customers:
\begin{itemize}
	\item[$\bullet$] $Add$ ($N_{Add}$): Adding one unselected customer to some position of some route. 
	\item[$\bullet$] $Drop$ ($N_{Drop}$): Removing one customer from one route. 
\end{itemize}

It is interesting to note that Pei et al. \cite{pei_solving_2020} introduced a series of data structures to realize fast evaluation of the neighboring solutions in the neighborhoods $N_1\mbox{-}N_4$, $N_{Add}$ and $N_{Drop}$ for solving the related TRPP. Here, we extend their method to the neighborhoods\footnote{The neighborhoods $N_5\mbox{-}N_7$ are not applied and studied in \cite{pei_solving_2020}.} for the MTRPP. In practice, evaluating each neighboring solution in our algorithm can be finished in $O(1)$, which is more efficient than the reference algorithms in the literature \cite{lu2019memetic, avci_adaptive_2019}.
The detailed comparisons of complexities for exploring different neighborhoods between the reference algorithms and the proposed algorithm are discussed in Section~\ref{subsec:Algo:Novel}.
The complexities of exploring the aforementioned neighborhoods are summarized as follows. 

\begin{mypro}\label{pro:1}
For the first set of four neighborhoods ($N_1\mbox{-}N_4$) and the third set of two neighborhoods ($N_{Add}$ and $N_{Drop}$) for the MTRPP, the complexity of evaluating each neighboring solution is $O(1)$. Let $n$ be the number of all customers and $m$ be the number of visited customers in the solution. The time complexities of exploring these neighborhoods are given as follows.
\begin{itemize}
	\item[a)] Exploring the complete $Swap$ neighborhood requires $O(m^2)$. 
	\item[b)] Exploring the complete $Insert$ neighborhood requires $O(m^2)$.
	\item[c)] Exploring the complete $2\mbox{-}opt$ neighborhood requires $O(m^2)$.
	\item[d)] Exploring the complete $Or\mbox{-}opt$ neighborhood requires $O(m^2\cdot h)$.
	\item[e)] Exploring the complete $Add$ neighborhood requires $O(m\cdot (n-m))$.
	\item[f)] Exploring the complete $Drop$ neighborhood requires $O(m)$.
\end{itemize}
\end{mypro}

\begin{mypro}\label{pro:2}
For the second set of three neighborhoods ($N_5\mbox{-}N_7$), evaluating each neighboring solution can be done in $O(1)$. Let m be the number of visited customers in the solution. The time complexities of exploring these neighborhoods are summarized as follows.
\begin{itemize}
	\item[a)] Exploring the complete $Inter\mbox{-}Swap$ neighborhood can be finished in $O(m^2)$.
	\item[b)] Exploring the complete $Inter\mbox{-}Insert$ neighborhood can be finished in $O(m^2)$.
	\item[c)] Exploring the complete $Inter\mbox{-}2\mbox{-}opt$ neighborhood can be finished in $O(m^2)$.	
\end{itemize}
\end{mypro}

Detailed proofs of Propositions~\ref{pro:1} and \ref{pro:2} are presented in Appendix~\ref{app1:proof}.
With Equation~(\ref{eq:fast}) and a special array in Equation~(\ref{eq:data}), we can efficiently explore the aforementioned neighborhoods.  
It is worth noting that, the collected revenues $p_i-l(i)$ of some customers may be negative during the search process while only non-negative profits are taken into consideration by Equation~(\ref{eq:fast}). Therefore, a local optimization based on the $Drop$ operator (line 13 in Algorithm~\ref{Algo:VNS}) is performed after other neighborhoods to get rid of this difficulty.

\subsection{Perturbation procedure}\label{subsec:Algo:Sperb}

In order to help the search escape from deep local optimum, we apply two operators $Insert$ and $Add$ to perturb the local optimum. We firstly perform $St$ times the $Insert$ operation by randomly choosing a route and inserting some customer to a random position in the route. For the $Add$ operation, we randomly add an unvisited customer to the tail of a random route. And this procedure repeats until all the unvisited customers are added to the solution. The parameter $St$ is determined by the experiments in Section~\ref{subsec:Experi:setup}. We also tested other perturbation methods but the proposed method proves to be the best.

\subsection{Arc-based crossover} \label{subsec:Algo:abx}

Memetic algorithms employ crossovers to generate diversified offspring solutions from parent solutions at each generation.
Generally, a meaningful crossover is expected to be able to inherit useful attributes of the parent solutions and maintain some diversity with respect to the parents \cite{hao2012memetic}. 

Preliminary experiments showed that the same arcs frequently appear in high-quality solutions (See Section~\ref{subsec:Ana:rational}), which naturally encourages us to preserve these shared arcs (meaningful components) in the offspring solution\footnote{Note that the idea of preserving and transferring the shared components from the parents to the offspring is the basis of backbone based crossovers \cite{zhang2004configuration, wang2013backbone}.}.  Following this observation, we propose a dedicated arc-based crossover for the MTRPP.

For a given solution $\varphi$ with $K$ paths $\{X_1, ..., X_K\}$, where each path $X_k=(x_0^k, ..., x_{m_k}^k )$ contains $m_k$ customers, the corresponding arc set $A$ is defined as follows:

\begin{equation}
A=\{(x_i^k,x_{i+1}^k): x_i^k, x_{i+1}^k\in X_k, i\in[0, m_k-1], k\in [1,K] \}	
\end{equation}

Given two parent solutions $\varphi_s$ and $\varphi_t$, let $V_s$ and $V_t$ represent 
the set of selected customers respectively, and $A_s$ and $A_t$ represent their corresponding arc sets. 
The arc-based crossover first copies one parent solution (say $\varphi_s$) to the offspring solution, then randomly inserts fifty percent of non-shared arcs of $\varphi_t$  ($A_t\setminus A_s$) into the offspring solution, and finally removes the duplicated vertices 
if needed.

%
%

\begin{algorithm}[!tb]
	\begin{small}
		\caption{The arc-based crossover (\emph{ABX})}
		\label{Algo:ABX}
		\begin{algorithmic}[1]
			\renewcommand{\algorithmiccomment}[1]{\hfill\textrm{// \small{#1}}}
			\STATE \sf \textbf{Input}: Input graph $G(V,E)$, parent solutions $\varphi_s$ and $\varphi_t$
			\STATE \textbf{Output}: Offspring solution $\varphi_o$
			\STATE /* RandomSel-Half-Arcs randomly selects 50\% of arcs from a given set of arcs. */
			\STATE /* $V_o$ is the set of the selected customers for $\varphi_o$. */
			\STATE /* $V_f$ is the set of nodes, which will be not removed or inserted to other positions in future operations. */
			
			\STATE $\varphi_o\leftarrow \varphi_s$
			\STATE $V_o\leftarrow V_s$
			\STATE $V_{f}=\emptyset$
			\STATE $A_u\leftarrow$  RandomSel-Half-Arcs($A_t\setminus A_s$)
			\FOR {Each arc $(a,b)\in A_s\cap A_t$}
			\STATE $V_{f}\leftarrow V_{f}\cup \{a\}$
			\STATE $V_{f}\leftarrow V_{f}\cup \{b\}$
			\ENDFOR
			
			\FOR {Each arc $(a,b)\in A_u$}
			\IF{$a\notin V_{o}$ and $b\notin V_{o}$} 
				\STATE Insert $(a,b)$ to the tail of some route in $\varphi_o$
				\STATE $V_{o}\leftarrow V_{o}\cup \{a\}$
				\STATE $V_{o}\leftarrow V_{o}\cup \{b\}$
		
			\ELSIF {$a\in V_{o}$ and $b\notin V_{o}$}
				\STATE Insert $b$ to the position after $a$ in $\varphi_o$
				\STATE $V_{o}\leftarrow V_{o}\cup \{b\}$ 	
			
			\ELSIF {$a\notin V_{o}$ and $b\in V_{o}$}
				\STATE Insert $a$ to the position before $b$ in $\varphi_o$
				\STATE $V_{o}\leftarrow V_{o}\cup \{a\}$
			
			\ELSIF{$b\notin V_f$ }		
				\STATE Remove $b$ from $\varphi_o$
				\STATE Insert $b$ to the position after $a$ in $\varphi_o$ 
						
			\ELSIF{$a\notin V_f$ and $b\in V_f$ }		
				\STATE Remove $a$ from $\varphi_o$
				\STATE Insert $a$ to the position before $b$ in $\varphi_o$ 
				
			\ENDIF
			
			\STATE $V_{f}\leftarrow V_{f}\cup \{a\}$
			\STATE $V_{f}\leftarrow V_{f}\cup \{b\}$	

			\ENDFOR
			
			\RETURN $ \varphi_o$
		\end{algorithmic}
	\end{small}
\end{algorithm}
%
%


The proposed \emph{ABX} crossover is presented in Algorithm~\ref{Algo:ABX}. It starts by copying $\varphi_s$ to $\varphi_o$, copying $V_s$ to $V_o$,  initializing $V_f$ as empty and generating an arc set $A_u$ by randomly selecting 50\% arcs from $A_t\setminus A_s$ (lines 6-9). To preserve the shared arcs $(a,b)\in A_s \cap A_t$ in the offspring solution $\varphi_o$, we then add the vertices of these arcs into the set $V_f$ (lines 10-13). The vertices
 in $V_f$ will 
not be considered in the future operations. After that, we insert each arc $(a,b)\in A_u$ into the offspring solution and remove the duplicated vertices (lines 14-34) according to the following conditions 
\begin{itemize}
	\item[1)] If both $a$ and $b$ are not included in $V_o$, the arc $(a,b)$ is added to the tail of some route and the two nodes are added into $V_o$ (lines 15-18).
	\item[2)] If only $a$ is contained in $V_o$, the node $b$ is inserted at the position after $a$ in $\varphi_o$ and is added into $V_o$ (lines 19-21).
	\item[3)] If only $b$ belongs to $V_o$, the node $a$ is inserted at the position before $b$ in $\varphi_o$ and is added into $V_o$ (lines 22-24).
	\item[4)] If the two nodes are already in $V_o$ and $b$ is not in $V_f$, we remove $b$ from $\varphi_o$ and insert it at the position after $a$ (lines 25-27).
	\item[5)] If the two nodes are already in $V_o$ and $a$ is not in $V_f$, we remove $a$ from  $\varphi_o$ and insert it at the position before $b$ (lines 28-31).
\end{itemize}
Both $a$ and $b$ are added into the set $V_f$ after the aforementioned operations (lines 32-33) and the whole loop ends when all the arcs $(a,b)\in A_u$ is added into the offspring. At last, the new generated offspring $\varphi_o$ is returned (line 35).

\begin{figure}[htbp]
	\centering
	\includegraphics[scale=0.35]{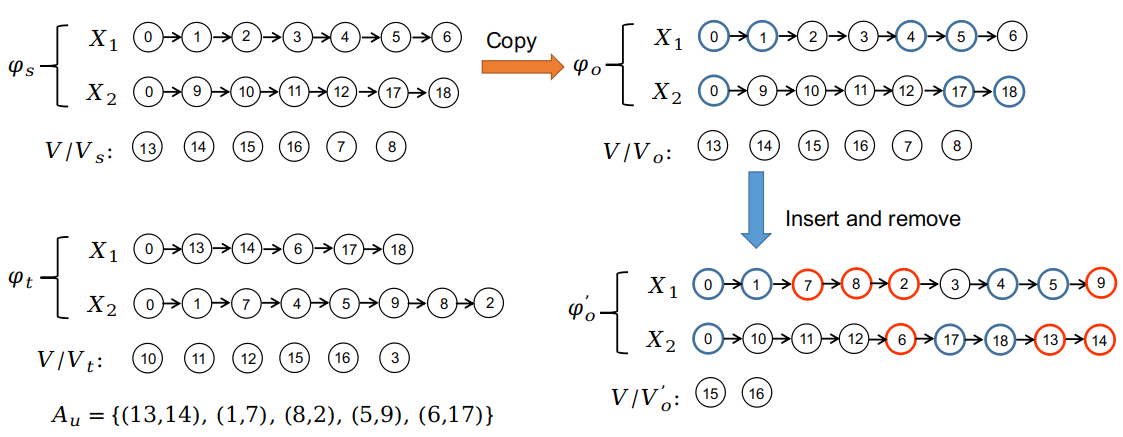}
	\caption{Illustration of \emph{ABX} on a 18-customer instance with two routes. 
	$\varphi_s$ and $\varphi_t$ are two parent solutions, $\varphi_o$ is the solution copied from $\varphi_s$ and $\varphi_o^\prime$ is the generated offspring solution. $A_u$ is the set of arcs, which are randomly selected from the non-shared arcs of $\varphi_t$. $V\setminus V_i$ represents the set of unselected customers for solution $\varphi_i$, which can be $\varphi_s$, $\varphi_t$, $\varphi_o$ and $\varphi_o^\prime$.
	The nodes of the shared arcs between $\varphi_s$ and $\varphi_t$ are marked in blue while the nodes involved in inserting and removing are marked in red.}
	\label{fig:abx}
\end{figure}

Figure~\ref{fig:abx} shows an illustrative example of the proposed crossover.

\subsection{Pool updating} \label{subsec:Algo:updating}

After the improvement of the offspring solution by the local refinement procedure, the population is updated by the improved offspring solution $\varphi_{lb}$ (See line 28 in Algorithm~\ref{Algo:EHSA-MTRPP}). In this work, we employ a simple strategy: if $\varphi_{lb}$ is different from all the solutions in the population and better than some solution in terms of the objective value, $\varphi_{lb}$ replaces the worst solution in the population. Otherwise, $\varphi_{lb}$ will be abandoned. We also tested other population updating strategies like diversification-quality strategy\footnote{This strategy considers not only the quality of the newly obtained solution but also its average distance to the other solutions to determine whether accepting $\varphi_{lb}$ into the population.}. However, this simple updating strategy has a better performance.

\subsection{Discussion}\label{subsec:Algo:Novel}

EHSA-MTRPP distinguishes itself from the reference algorithms \cite{lu2019memetic, avci_adaptive_2019} in two aspects.


EHSA-MTRPP employs fast neighborhood evaluation techniques in its local optimization procedure for the MTRPP for the first time. These evaluation techniques ensures a higher computational efficiency of neighborhood examination compared to the existing algorithms ALNS-MTRPP \cite{avci_adaptive_2019} and MA-MTRPP \cite{lu2019memetic}. 
To illustrate this point, Table~\ref{tab:sum:neighborhood} summarizes the different neighborhoods as well as the complexities to explore each neighborhood in ALNS-MTRPP and MA-MTRPP. 
\begin{table}[htbp]
\centering
\caption{Summary of the neighborhood structures as well as their complexities in the reference algorithms and the proposed algorithm, where $n$ depicts the number of customers, $m$ is the number of the selected customers and $h$ is the number of consecutive customers in the block for $N_4$.}
\label{tab:sum:neighborhood}
\renewcommand{\arraystretch}{1.6}
\setlength{\tabcolsep}{1.3mm}
\begin{scriptsize}
\begin{tabular}{llcclcclcc}
\noalign{\smallskip}
\hline
\multirow{2}{*}{Neighborhood} & & \multicolumn{2}{c}{ALNS-MTRPP \cite{avci_adaptive_2019}} & & 
\multicolumn{2}{c}{MA-MTRPP \cite{lu2019memetic}} & & \multicolumn{2}{c}{EHSA-MTRPP} \\

\cline{3-4}
\cline{6-7}
\cline{9-10}

 & & Employment  & Complexity & & Employment  & Complexity & & Employment  & Complexity \\
\hline

$Swap$ 	&  & \Checkmark   &  $O(m^2\lg n)$	&  &\Checkmark   &  $O(m^3)$	 &  & \Checkmark   &  $O(m^2)$			\\

$Insert$ 	&  & \Checkmark   &  $O(m^2\lg n)$	&  &\Checkmark   &  $O(m^3)$	 &  & \Checkmark   &  $O(m^2)$				\\

$2\mbox{-}opt$ 	&  & \Checkmark   &  $O(m^2\lg n)$	&  &\Checkmark   &   $O(m^3)$	 &  & \Checkmark   &  $O(m^2)$				\\

$Or\mbox{-}opt$ 	&  & \Checkmark   &  $O(m^2\lg n)$	&  &\Checkmark   &   $O(m^3)$	 &  & \Checkmark   &  $O(m^2\cdot h)$				\\

$Inter\mbox{-}Swap$ 	&  & \Checkmark   & $O(m^2\lg n)$	&  &\Checkmark   &   $O(m^3)$	 &  & \Checkmark   &  $O(m^2)$				\\

$Inter\mbox{-}Insert$ 	&  & \Checkmark   &  $O(m^2\lg n)$	&  &\Checkmark   &   $O(m^3)$	 &  & \Checkmark   &  $O(m^2)$				\\

$Inter\mbox{-}2\mbox{-}opt$ 	&  & \Checkmark   &  $O(m^2\lg n)$	&  &\Checkmark   &   $O(m^3)$	 &  & \Checkmark   &  $O(m^2)$				\\

$Inter\mbox{-}Or\mbox{-}opt$ 	&  & \Checkmark   &  $O(m^2\lg n)$	&  &\Checkmark   &   $O(m^3)$	 &  & \XSolid   &  -				\\

$Add$ 	&  & \XSolid   &  -	&  &\XSolid   &  -	 &  & \Checkmark   &  $O(m\cdot (n-m))$				\\

$Drop$ 	&  & \XSolid   &  -	&  &\XSolid   &  -	 &  & \Checkmark   &  $O(m)$			\\

$Double\mbox{-}bridge$ 	&  & \XSolid   &  -	&  &\Checkmark   &   $O(m^3)$	 &  & \XSolid   &  -			\\

\hline
\end{tabular}
\end{scriptsize}
\label{tab:parameter}
\end{table}


From Table~\ref{tab:sum:neighborhood}, we clearly remark that the proposed algorithm explores more efficiently the used neighborhoods. It is worth noting that we also proved that evaluating one neighboring solution in the $Inter\mbox{-}Or\mbox{-}opt$ and $Double\mbox{-}bridge$\footnote{$Double\mbox{-}bridge$ is a popular operator for the traveling salesman problem \cite{lin1973effective}.} based neighborhoods can be done in $O(1)$ with the help of Equation~(\ref{eq:fast}) and the auxiliary data structures. However, we observed that these neighborhoods are not helpful in improving the performance of our algorithm. Therefore, these two neighborhoods are not employed in the proposed algorithm.

Additionally, 
the proposed algorithm adopts a dedicated arc-based crossover, which is able to generate new offspring solutions inheriting meaningful components (the shared arcs) and getting diversified from the parent solutions.
MA-MTRPP \cite{lu2019memetic} applies a route-based crossover (RBX) \cite{potvin1996vehicle}, which simply copies one parent solution to the offspring solution, replaces some route of the offspring solution with a route from another parent solution and removes the duplicated vertices if needed. 
Our experiments and observations showed that the key components of the solutions are the `arcs' but not the `routes', making ABX more appropriate than RBX for solving the MTRPP. 
Experimental results in Section~\ref{subsec:Analyse:rbx} confirm these observations and demonstrate the effectiveness of the proposed algorithm with ABX compared to its variant with RBX.


\section{Computational results and comparative study}\label{sec:Experi}

This section presents computational experiments over the benchmark instances in the literature to evaluate the  EHSA-MTRPP algorithm.  

\subsection{Instances, reference algorithms and parameter setting} \label{subsec:Experi:setup}

Our computational experiments are based on two groups of 470 benchmark instances\footnote{The instances are collected from the authors of \cite{avci_adaptive_2019,lu2019memetic} and can be download from \url{https://github.com/REN-Jintong/MTRPP}.}, including 230 instances proposed by Avci and Avci \cite{avci_adaptive_2019} (denoted by Ins\_Avci) and 240 instances proposed by Lu et al. \cite{lu2019memetic} (denoted by Ins\_Lu).

%
%

%
%
The 230 Ins\_Avci instances are divided into 14 sets of instances according to the number of customers and servers (repairmen). The first ten sets of instances are converted from instances of the TRPP \cite{dewilde2013heuristics} ($n$=10, 20, 50, 100, 200) by considering two and three servers. As each set of instances of the TRPP is composed of 20 instances, the authors have created $5\times 2 \times 20=200$ instances. The other four sets of instances are of larger sizes, including 10 instances with 500 customers and 10 servers, ten instances with 500 customers and 20 servers, five instances with 750 customers and 100 servers, and five instances with 1000 customers and 50 servers. 

The 240 Ins\_Lu instances are also based on instances of the TRPP \cite{dewilde2013heuristics}, and divided into 12 sets of instances with $n$=20, 50, 100, 200 and two, three and four servers. Unlike the Ins\_Avci instances, Lu et al. adjusted the profit for each customer of the instances in order to ensure a high-quality solution to hold about 75\% to 95\% of all the customers. For this purpose, each customer $i$ is assigned a non-negative profit $p_i$, a random integer between $\lceil d_{0, i} \rceil$ and $\lceil \frac{n}{k} \times \frac{\sum_{(i,j)\in E} d_{i,j}}{|E|} \rceil$, where $n$ is the number of all customers and $k$ is the number of servers.

The proposed EHSA-MTRPP algorithm was programmed in C++ and compiled with the g++ 7.5.0 compiler and -O3 optimization flag\footnote{The source code will be made available at \url{https://github.com/REN-Jintong/MTRPP} upon the publication of this work.}. The experiments were carried out on a computer with an Intel Xeon(R) E5-2695 processor (2.1 GHz CPU and 2 GB RAM), which is slower than the 2.4GHz computers used in the literature \cite{avci_adaptive_2019, lu2019memetic}.

The reference results in the literature were reported in \cite{avci_adaptive_2019} for the 230 Ins\_Avci instances with the ALNS-MTRPP algorithm and \cite{lu2019memetic} for the 240 Ins\_Lu instances with the MA-MTRPP algorithm. Unfortunately, the source codes of these reference algorithms are not available. Therefore, to ensure a fair comparison, we performed two different experiments on the two groups of instances. Following the experimental setup in \cite{avci_adaptive_2019, lu2019memetic}, EHSA-MTRPP was run independently 5 times with different seeds on each Ins\_Avci instance while 10 times on the Ins\_Lu instances. The stopping condition in the literature is a prefixed maximum number of iterations, while EHSA-MTRPP uses a maximum cutoff time. One notes that the average running time of ALNS-MTRPP \cite{avci_adaptive_2019} usually reaches several hours for the instances of large size and the average time to get the best found solution for MA-MTRPP \cite{lu2019memetic} is around 300 and 500 seconds for the 200-customer instances. For fair comparisons, we set our cutoff time $T_{max}$ to be twice the number of customers (in seconds). In practice, EHSA-MTRPP is able to attain better solutions than the reference algorithms with less time for most of  the benchmark instances.


To determine the parameters listed in Table~\ref{tab:input}, we employed an automatic parameter tuning tool Irace \cite{lopez2016irace}. In this experiment, we selected 10 large and difficult instances as the training instances and set the maximum number of runs (tuning budget) to 2000.
\begin{table}[htbp]
\centering
\caption{Parameters of EHSA-MTRPP tuned with the Irace package.}
\label{tab:input}
\renewcommand{\arraystretch}{1.6}
\setlength{\tabcolsep}{2.8mm}
\begin{scriptsize}
\begin{tabular}{lp{0.35\columnwidth}cc}
\noalign{\smallskip}
\hline
\multicolumn{1}{c}{Parameter} & \multicolumn{1}{c}{Description}  & \multicolumn{1}{c}{Type} & \multicolumn{1}{c}{Value range}\\
\hline

$Limi$ 	&     Search limit    				   			&  Integer	&	[0, 30]\\
$St$ 	&     Strength of the $Insert$ perturbation   	&  Integer	&	[0, 100]\\
$Nump$ 	&     Number of population    				   	&  Categorical	&	\{6, 8, 10, 20, 50, 100\}\\
\hline
\end{tabular}
\end{scriptsize}
\label{tab:parameter}
\end{table}
According to the tuning experiment, the parameters determined by Irace are: $Limi=2$, $St=11$ and $Nump=10$.
We used this parameter setting for EHSA-MTRPP for all our computational experiments.

\subsection{Comparative studies}\label{subsec:Experi:com}

This section presents the experimental results obtained by EHSA-MTRPP with respect to the reference algorithms \cite{lu2019memetic,avci_adaptive_2019} over the two groups of 470 benchmark instances. 

Table~\ref{tab:allresults:avci} lists the overall results of the reference algorithm ALNS-MTRPP and our EHSA-MTRPP algorithm over the 230 Ins\_Avci instances (better results are indicated in bold). Column `Problem set' shows the instances with different numbers of customers (Size) and servers (k). Column `UB' represents the best-known upper bounds of the instances
\cite{avci_adaptive_2019, lu2019memetic} and column `Bestsofar' describes the best found solutions (lower bounds) in the literature. Columns `Best', `Average' and `Tavg' (columns 4-6) show respectively the best found results, the average found results and the average time to obtain the best found solutions for ALNS-MTRPP. The following three columns depict the same information for EHSA-MTRPP (all the values aforementioned are averaged over the instances of each set). Column `p-value' lists the results of the Wilcoxon signed rank tests of the best found results (column `Best') between ALNS-MTRPP and EHSA-MTRPP, where `NA' indicates no difference between the two groups of results. Then column `$\delta$' presents the improvement in percentage of the best objective value found by EHSA-MTRPP over the best objective value of ALNS-MTRPP. The last three columns list the number of instances for which our EHSA-MTRPP algorithm improved (`Win'), matched (`Match') or failed (`Fail') to attain the best found results reported in \cite{avci_adaptive_2019}. Finally, row `Avg.' depicts the average values of the corresponding indicators for all 230 Ins\_Avci instances.


\begin{table}[htbp]
\centering
\renewcommand{\arraystretch}{1.5}
\setlength{\tabcolsep}{0.5mm}
\caption{Results of the reference algorithm ALNS-MTRPP \cite{avci_adaptive_2019} and EHSA-MTRPP on the 230 Ins\_Avci instances. Each instance was solved 5 times according to \cite{avci_adaptive_2019}. The data in column `Bestsofar' are the compiled results from ALNS-MTRPP and GRASP-ILS \cite{avci_adaptive_2019}. The optimal solutions for the instances of `Size=10, k=2 and k=3' are known but their timing information is not available.}{\smallskip}\label{tab:allresults:avci}

\resizebox{\textwidth}{35mm}{

\begin{tabular}{lllclllclllclllll}
\toprule
\multirow{2}{*}{Problem set}& \multirow{2}{*}{UB \cite{avci_adaptive_2019}}& \multirow{2}{*}{Bestsofar\cite{avci_adaptive_2019}}  & & \multicolumn{3}{c}{ALNS-MTRPP} & & \multicolumn{3}{c}{EHSA-MTRPP} & & \multirow{2}{*}{p-value} & \multirow{2}{*}{$\delta$} & \multirow{2}{*}{W}& \multirow{2}{*}{M}& \multirow{2}{*}{F} \\
\cline{5-7}
\cline{9-11}

 & & &  &\multicolumn{1}{l}{Best}& \multicolumn{1}{l}{Average} & \multicolumn{1}{l}{Tavg} & & \multicolumn{1}{l}{Best} &\multicolumn{1}{l}{Average} & \multicolumn{1}{l}{Tavg} & & &\\  

\midrule
Size=10, m=2     &  2114.85 &  2114.85 &  &			 2114.85 &  2114.85 &     0.00& &			 2114.85 &  2114.85 &     0.03& &			NA       & 0.000000\% &			0     & 20    & 0     \\ 
Size=10, m=3     &  2230.60 &  2230.60 &  &			 2230.60 &  2230.60 &     0.00& &			 2230.60 &  2230.60 &     0.03& &			NA       & 0.000000\% &			0     & 20    & 0     \\ 
Size=20, m=2     &  9680.85 &  9074.60 &  &			 9074.60 &  9074.60 &     3.15& &			 9074.60 &  9074.60 &     0.06& &			NA       & 0.000000\% &			0     & 20    & 0     \\ 
Size=20, m=3     &  9994.85 &  9450.45 &  &			 9450.45 &  9450.45 &     3.10& &			 9450.45 &  9450.45 &     0.06& &			NA       & 0.000000\% &			0     & 20    & 0     \\ 
Size=50, m=2     & 57587.75 & 55469.15 &  &			55469.15 & 55468.55 &    35.50& &			55469.15 & 55469.15 &     0.82& &			NA       & 0.000000\% &			0     & 20    & 0     \\ 
Size=50, m=3     & 58821.75 & 57184.85 &  &			57184.85 & 57184.45 &    30.85& &			\textbf{57185.35} & \textbf{57185.35} &     0.78& &			3.17$\times 10^{-1}$ & 0.000874\% &			1     & 19    & 0     \\ 
Size=100, m=2     & 232351.15 & 226900.95 &  &			226899.95 & 226895.80 &   346.45& &			\textbf{226900.95} & \textbf{226900.47} &    22.96& &			1.02$\times 10^{-1}$ & 0.000441\% &			0     & 20    & 0     \\ 
Size=100, m=3     & 235956.05 & 231957.70 &  &			231954.05 & 231947.30 &   551.05& &			\textbf{231958.70} & \textbf{231954.23} &    29.80& &			1.80$\times 10^{-1}$ & 0.002005\% &			1     & 19    & 0     \\ 
Size=200, m=2     & 907250.15 & 893197.90 &  &			893183.35 & 892864.45 &  3600.00& &			\textbf{893513.85} & \textbf{893374.88} &   263.23& &			1.20$\times 10^{-4}$ & 0.037002\% &			19    & 0     & 1     \\ 
Size=200, m=3     & 917633.35 & 907775.35 &  &			907775.35 & 907611.55 &  3600.00& &			\textbf{907950.35} & \textbf{907841.50} &   258.94& &			8.84$\times 10^{-5}$ & 0.019278\% &			20    & 0     & 0     \\ 
Size=500, m=10    & 1523086.90 & 1428729.50 &  &			1428716.30 & 1422361.10 & 10800.00& &			\textbf{1437256.40} & \textbf{1436265.76} &   898.97& &			5.06$\times 10^{-3}$ & 0.597746\% &			10    & 0     & 0     \\ 
Size=500, m=20    & 766209.10 & 692225.50 &  &			692074.30 & 688804.60 & 10800.00& &			\textbf{694406.60} & \textbf{694114.40} &   897.70& &			5.06$\times 10^{-3}$ & 0.337001\% &			10    & 0     & 0     \\ 
Size=750, m=100   & 4150788.40 & 4000423.40 &  &			4000199.00 & 3966184.40 & 43200.00& &			\textbf{4000585.60} & \textbf{4000541.60} &  1352.60& &			4.31$\times 10^{-2}$ & 0.009665\% &			5     & 0     & 0     \\ 
Size=1000, m=50    & 5402658.80 & 5189124.00 &  &			5186645.80 & 5066567.40 & 43200.00& &			\textbf{5191726.40} & \textbf{5191527.76} &  1757.60& &			4.31$\times 10^{-2}$ & 0.097955\% &			5     & 0     & 0     \\ 
\hline
Avg. & 518837.49 & 500280.07 &  &		500212.50 & 496401.17 &  3527.83& &		\textbf{500848.55} & \textbf{500765.52} &   195.88& &		 &  &		 &  &  \\ 
\hline
Sum &  &  &  & &  &  & & &  &  & &		 &  &		 71 & 158 &  1\\ 
\bottomrule
\end{tabular}
}
\end{table}
From row `Avg.' of Table~\ref{tab:allresults:avci}, we remark that EHSA-MTRPP outperforms ALNS-MTRPP concerning the best found results and the average found results.
The two algorithms have the same performance for the first five sets of instances (instances of small sizes), while for the remaining nine sets of instances, EHSA-MTRPP performs better than the reference algorithm ALNS-MTRPP both in terms of solution quality (`Best' and `Average') and running time (column `Tavg').
In particular, the results of the Wilcoxon signed rank test (column `p-value') show that there exists a significant difference of the best found results between ALNS-MTRPP and EHSA-MTRPP over the last six set of instances (p-value$<$0.05).
Overall, EHSA-MTRPP clearly dominates ALNS-MTRPP by updating the best records (new lower bounds) for 71 instances, matching the best-known results for 158 instances and only missing one best-known result.

Using similar column headings as Table~\ref{tab:allresults:avci} (excluding column `Bestsofar'),  Table~\ref{tab:allresults:lu} summarizes the overall results of MA-MTRPP and EHSA-MTRPP on the 240 Ins\_Lu instances. 
\begin{table}[htbp]
\centering
\renewcommand{\arraystretch}{1.5}
\setlength{\tabcolsep}{0.5mm}
\caption{Results of the reference algorithm MA-MTRPP \cite{lu2019memetic} and  EHSA-MTRPP on the instances of Ins\_Lu. Each instance was solved 10 times according to \cite{lu2019memetic}. The optimal solutions for the instances of small size (`Size=20, k=2, 3 and 4') are known.}{\smallskip}\label{tab:allresults:lu}
\begin{tiny}

\begin{tabular}{llclllclllclllll}
\toprule
\multirow{2}{*}{Problem set}& \multirow{2}{*}{UB \cite{lu2019memetic}}  & & \multicolumn{3}{c}{MA-MTRPP} & & \multicolumn{3}{c}{EHSA-MTRPP} & & \multirow{2}{*}{p-value} & \multirow{2}{*}{$\delta$} & \multirow{2}{*}{W}& \multirow{2}{*}{M}& \multirow{2}{*}{F} \\
\cline{4-6}
\cline{8-10}

 & &   &\multicolumn{1}{l}{Best}& \multicolumn{1}{l}{Average} & \multicolumn{1}{l}{Tavg} & & \multicolumn{1}{l}{Best} &\multicolumn{1}{l}{Average} & \multicolumn{1}{l}{Tavg} & & &\\  

\midrule

Size=20, m=2     &  3937.60 &  &		 3937.60 &  3937.60 &     1.31& &		 3937.60 &  3937.60 &     0.05& &		NA       & 0.000000\% &		0     & 20    & 0     \\ 
Size=20, m=3     &  2399.20 &  &		 2399.20 &  2399.20 &     1.29& &		 2399.20 &  2399.20 &     0.05& &		NA       & 0.000000\% &		0     & 20    & 0     \\ 
Size=20, m=4     &  1733.40 &  &		 1733.40 &  1733.40 &     1.23& &		 1733.40 &  1733.40 &     0.06& &		NA       & 0.000000\% &		0     & 20    & 0     \\ 
Size=50, m=2     & 29677.45 &  &			27172.15 & 27172.15 &     7.38& &			\textbf{27173.55} & \textbf{27173.55} &     1.02& &			1.09$\times 10^{-1}$ & 0.005152\% &			3     & 17    & 0     \\ 
Size=50, m=3     & 19464.60 &  &		17523.55 & 17523.55 &     6.26& &		17523.55 & 17523.55 &     0.66& &		NA       & 0.000000\% &		0     & 20    & 0     \\ 
Size=50, m=4     & 14805.10 &  &			13049.05 & 13049.05 &     5.72& &			\textbf{13049.25} & \textbf{13049.25} &     0.75& &			1.80$\times 10^{-1}$ & 0.001533\% &			2     & 18    & 0     \\ 
Size=100, m=2     & 120082.50 &  &			113566.35 & 113560.76 &    46.13& &			\textbf{113567.10} & \textbf{113566.60} &    21.44& &			1.80$\times 10^{-1}$ & 0.000660\% &			2     & 18    & 0     \\ 
Size=100, m=3     & 81703.80 &  &			76976.35 & 76972.48 &    37.85& &			\textbf{76976.65} & \textbf{76976.24} &    23.31& &			3.17$\times 10^{-1}$ & 0.000390\% &			1     & 19    & 0     \\ 
Size=100, m=4     & 61265.85 &  &			57188.40 & 57186.69 &    32.55& &			\textbf{57188.55} & \textbf{57188.53} &    21.07& &			3.17$\times 10^{-1}$ & 0.000262\% &			1     & 19    & 0     \\ 
Size=200, m=2     & 489139.40 &  &			472301.40 & 472002.08 &   455.39& &			\textbf{472499.25} & \textbf{472354.94} &   254.16& &			1.03$\times 10^{-4}$ & 0.041891\% &			19    & 0     & 1     \\ 
Size=200, m=3     & 333141.60 &  &			321136.55 & 320912.21 &   358.27& &			\textbf{321278.75} & \textbf{321175.57} &   245.66& &			8.86$\times 10^{-5}$ & 0.044280\% &			20    & 0     & 0     \\ 
Size=200, m=4     & 246431.05 &  &			236694.15 & 236539.09 &   278.37& &			\textbf{236805.20} & \textbf{236720.93} &   229.22& &			1.55$\times 10^{-4}$ & 0.046917\% &			18    & 1     & 1     \\ 
\hline
Avg. & 116981.80 &  &		111973.18 & 111915.69 &   102.64& &		\textbf{112011.00} & \textbf{111983.28} &    66.45& &		 &  &		 &  &  \\ 

\hline
Sum &  &  &  & &  &  & & &  &  & 		 &  &		 66 & 172 &  2\\

\bottomrule
\end{tabular}
\end{tiny}
\end{table}
From row `Avg.' of Table~\ref{tab:allresults:lu}, one observes that EHSA-MTRPP achieves a better performance (column `Best' and `Average') than MA-MTRPP with a shorter average time (66.45 seconds for EHSA-MTRPP vs 102.64 seconds for MA-MTRPP).
As to each set of instances, EHSA-MTRPP shows a better or equal performance concerning the best found results and average found results. In particular, the proposed algorithm outperforms MA-MTRPP on the last three sets of large instances confirmed by the Wilcoxon signed rank test (p-value $<$ 0.05).
In addition, EHSA-MTRPP spends less time (column `Tavg') than MA-MTRPP to attain the best found solutions for each set of instances.
Overall, EHSA-MTRPP updates 66 best records (new lower bounds), matches the best-known results for 172 instances and misses only two best-known results.

In summary, EHSA-MTRPP provides much better results than the reference algorithms on the 470 benchmark instances by establishing 137 new record results (29\%) and matching best-known results for 330 instances (70\%).
The detailed comparisons between the reference algorithms \cite{lu2019memetic,avci_adaptive_2019} and EHSA-MTRPP are presented in Appendix~\ref{sec:Appendix}.


\section{Additional results}\label{sec:Analyse}

This section firstly presents the additional results to demonstrate the important roles of the fast evaluation technique and the arc-based crossover for the proposed algorithm. Furthermore, we experimentally compare the proposed algorithm to the variant with RBX, and reveal the rationale behind the proposed crossover.

\subsection{Influence of the fast evaluation technique in the neighborhood structure.}\label{subsec:Analyse:fast}

To investigate the influence of the fast evaluation technique on our algorithm, we create a variant of EHSA-MTRPP where the fast evaluation technique (named EHSA-MTRPP-NoFast) is disabled. We employ the parameters in Section~\ref{subsec:Experi:setup} and run both EHSA-MTRPP-NoFast and EHSA-MTRPP 10 independently times on each instance of large size ($n\geq 200$). The cut-off time is also set to be twice the number of customers.  Using similar column headings as Table~\ref{tab:allresults:lu}, Table~\ref{tab:nofast} gives the comparative results of EHSA-MTRPP-NoFast and EHSA-MTRPP over the instances of large size from Ins\_Avci and Ins\_Lu. (Better results are marked in bold.)
\begin{table}[htbp]
\centering
\renewcommand{\arraystretch}{1.5}
\setlength{\tabcolsep}{0.4mm}
\caption{Results of EHSA-MTRPP-NoFast and EHSA-MTRPP on the benchmark instances of large size. Each instance is solved 10 times and the cutoff time is set to be twice the number of customers.}{\smallskip}\label{tab:nofast}
\begin{tiny}

\begin{tabular}{llclllclllclllll}
\toprule
\multirow{2}{*}{Problem set}& \multirow{2}{*}{UB}  & & \multicolumn{3}{c}{EHSA-MTRPP-NoFast} & & \multicolumn{3}{c}{EHSA-MTRPP} & & \multirow{2}{*}{p-value} & \multirow{2}{*}{$\delta$} & \multirow{2}{*}{W}& \multirow{2}{*}{M}& \multirow{2}{*}{F} \\
\cline{4-6}
\cline{8-10}

 & & &\multicolumn{1}{l}{Best}& \multicolumn{1}{l}{Average} & \multicolumn{1}{l}{Tavg} & & \multicolumn{1}{l}{Best} &\multicolumn{1}{l}{Average} & \multicolumn{1}{l}{Tavg} & & &\\  
\midrule

Ins\_Avci &  &  & &  & & & & & & & & &	&  &  \\ 
\hline 
Size=200, m=2     & 907250.15  &  &			893247.90 & 892891.18 &   169.58& &			\textbf{893513.85} & \textbf{893374.88} &   263.23& &			1.03$\times 10^{-4}$ & 0.029773\% &			19    & 0     & 1     \\ 
Size=200, m=3     & 917633.35  &  &			907732.95 & 907532.35 &   169.48& &			\textbf{907950.35} & \textbf{907841.50} &   258.94& &			8.86$\times 10^{-5}$ & 0.023950\% &			20    & 0     & 0     \\ 
Size=500, m=10    & 1523086.90  &  &			1431017.90 & 1428561.78 &   500.05& &			\textbf{1437256.40} & \textbf{1436265.76} &   898.97& &			5.06$\times 10^{-3}$ & 0.435948\% &			10    & 0     & 0     \\ 
Size=500, m=20    & 766209.10  &  &			692814.30 & 691839.16 &   500.21& &			\textbf{694406.60} & \textbf{694114.40} &   897.70& &			5.06$\times 10^{-3}$ & 0.229831\% &			10    & 0     & 0     \\ 
Size=750, m=100   & 4150788.40  &  &			3999139.60 & 3948547.72 &   754.74& &			\textbf{4000585.60} & \textbf{4000541.60} &  1352.60& &			4.31$\times 10^{-2}$ & 0.036158\% &			5     & 0     & 0     \\ 
Size=1000, m=50    & 5402658.80  &  &			5180266.20 & 5116501.24 &  1003.76& &			\textbf{5191726.40} & \textbf{5191527.76} &  1757.60& &			4.31$\times 10^{-2}$ & 0.221228\% &			5     & 0     & 0     \\ 
\midrule
Ins\_Lu &  &  & &  & & & & & & & & &	&  &  \\ 
\hline
Size=200, m=2     & 489139.40 &  &			472354.10 & 471965.12 &   325.62& &			\textbf{472499.25} & \textbf{472354.94} &   254.16& &			7.80$\times 10^{-4}$ & 0.030729\% &			19    & 0     & 1     \\ 
Size=200, m=3     & 333141.60 &  &			321165.95 & 320919.83 &   327.09& &			\textbf{321278.75} & \textbf{321175.57} &   245.66& &			1.32$\times 10^{-4}$ & 0.035122\% &			19    & 1     & 0     \\ 
Size=200, m=4     & 246431.05 &  &			236709.90 & 236544.92 &   330.46& &			\textbf{236805.20} & \textbf{236720.93} &   229.22& &			1.32$\times 10^{-4}$ & 0.040260\% &			19    & 1     & 0     \\ 

\bottomrule
\end{tabular}
\end{tiny}
\end{table}
From column `Best' and `Average' in Table~\ref{tab:nofast}, one can conclude that EHSA-MTRPP outperforms EHSA-MTRPP-NoFast for each set of instances, which is also confirmed by the Wilcoxon signed rank tests ($p$-value$<0.05$). 

\begin{figure}[htbp]
	\centering
	\includegraphics[scale=0.3]{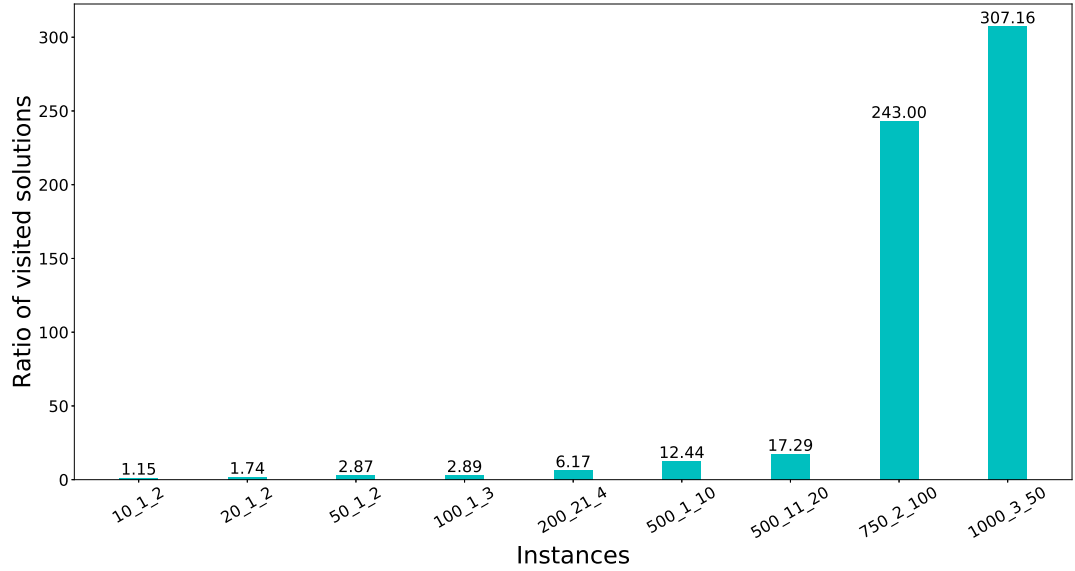}
	\caption{The average ratio of the visited solutions of EHSA-MTRPP over EHSA-MTRPP-NoFast for 9 instances of different sizes. The name `Size\_index\_k' for each instance indicates respectively the number of customers, the instance index and the number of routes. Each instance is solved 10 times independently and cut-off time is set to twice the number of customers.}
	\label{fig:bar}
\end{figure}

To illustrate the effectiveness of the fast evaluation technique, we carry out another experiment by running both algorithms independently 10 times on 9 instances of different sizes and record the numbers of the visited neighboring solutions. The cut-off time is also set to be twice the number of customers. Figure~\ref{fig:bar} shows the average ratio of the visited solutions of EHSA-MTRPP over EHSA-MTRPP-NoFast for the instances of different sizes. 
One observes that EHSA-MTRPP visits more neighboring solutions than EHSA-MTRPP-NoFast for all selected instances. The dominance of our algorithm with the fast evaluation technique becomes even more clear as the size of instance increases.
In summary, the results in Table~\ref{tab:nofast} and in Figure~\ref{fig:bar} demonstrate that the fast evaluation technique can help the proposed algorithm to explore more efficiently the search space and it indeed contributes to the performance of the proposed algorithm.

\subsection{Influence of the crossover operator}
\label{subsec:nox}

This section explores the contributions of the arc-based crossover to our algorithm. 
We create an EHSA-MTRPP variant (ILS-MTRPP) by disabling ABX (lines 15-16) in Algorithm~\ref{Algo:EHSA-MTRPP}. Using the same experimental setup in Section~\ref{subsec:Experi:setup}, another experiment is performed on the benchmark instances of large size, and the results are presented in Table~\ref{tab:nox} with the same column headings as Table~\ref{tab:nofast} (better results are marked in bold).

\begin{table}[htbp]
\centering
\renewcommand{\arraystretch}{1.5}
\setlength{\tabcolsep}{0.4mm}
\caption{Results of ILS-MTRPP and EHSA-MTRPP on the instances of large size from the benchmark. Each instance is solved 10 times and the cutoff time is set to be twice the number of customers.}{\smallskip}\label{tab:nox}
\begin{tiny}

\begin{tabular}{llclllclllclllll}
\toprule
\multirow{2}{*}{Problem set}& \multirow{2}{*}{UB}  & & \multicolumn{3}{c}{ILS-MTRPP} & & \multicolumn{3}{c}{EHSA-MTRPP} & & \multirow{2}{*}{p-value} & \multirow{2}{*}{$\delta$} & \multirow{2}{*}{W}& \multirow{2}{*}{M}& \multirow{2}{*}{F} \\
\cline{4-6}
\cline{8-10}

 & & &\multicolumn{1}{l}{Best}& \multicolumn{1}{l}{Average} & \multicolumn{1}{l}{Tavg} & & \multicolumn{1}{l}{Best} &\multicolumn{1}{l}{Average} & \multicolumn{1}{l}{Tavg} & & &\\  

\midrule
 
Ins\_Avci &  &  & &  & & & & & & & & &	&  &  \\ 
\hline 
Size=200, m=2     & 907250.15  &  &			893225.25 & 892997.62 &   106.16& &			\textbf{893513.85} & \textbf{893374.88} &   263.23& &			8.86$\times 10^{-5}$ & 0.032310\% &			20    & 0     & 0     \\ 
Size=200, m=3     & 917633.35  &  &			907796.10 & 907666.30 &    97.70& &			\textbf{907950.35} & \textbf{907841.50} &   258.94& &			8.84$\times 10^{-5}$ & 0.016992\% &			20    & 0     & 0     \\ 
Size=500, m=10    & 1523086.90  &  &			1435567.80 & 1434542.62 &   213.20& &			\textbf{1437256.40} & \textbf{1436265.76} &   898.97& &			5.06$\times 10^{-3}$ & 0.117626\% &			10    & 0     & 0     \\ 
Size=500, m=20    & 766209.10  &  &			693796.30 & 693456.04 &   215.12& &			\textbf{694406.60} & \textbf{694114.40} &   897.70& &			5.06$\times 10^{-3}$ & 0.087965\% &			10    & 0     & 0     \\ 
Size=750, m=100   & 4150788.40  &  &			4000456.40 & 4000414.68 &   320.81& &			\textbf{4000585.60} & \textbf{4000541.60} &  1352.60& &			4.31$\times 10^{-2}$ & 0.003230\% &			5     & 0     & 0     \\ 
Size=1000, m=50    & 5402658.80  &  &			5191550.60 & 5191326.92 &   520.71& &			\textbf{5191726.40} & \textbf{5191527.76} &  1757.60& &			4.31$\times 10^{-2}$ & 0.003386\% &			5     & 0     & 0     \\ 
\midrule
Ins\_Lu &  &  & &  & & & & & & & & &	&  &  \\ 
\hline
Size=200, m=2     & 489139.40 &  &			472282.15 & 472016.28 &   198.16& &			\textbf{472499.25} & \textbf{472354.94} &   254.16& &			2.93$\times 10^{-4}$ & 0.045968\% &			19    & 0     & 1     \\ 
Size=200, m=3     & 333141.60 &  &			321140.15 & 320982.75 &   201.67& &			\textbf{321278.75} & \textbf{321175.57} &   245.66& &			8.84$\times 10^{-5}$ & 0.043159\% &			20    & 0     & 0     \\ 
Size=200, m=4     & 246431.05 &  &			236703.05 & 236592.84 &   199.80& &			\textbf{236805.20} & \textbf{236720.93} &   229.22& &			8.84$\times 10^{-5}$ & 0.043155\% &			20    & 0     & 0     \\ 

\bottomrule
\end{tabular}
\end{tiny}
\end{table}

From Table~\ref{tab:nox}, one concludes that EHSA-MTRPP clearly dominates ILS-MTRPP in terms of columns `Best' and `Average' for each set of instances (only missing one instance in `Size=200, m=2' in Ins\_Lu). The dominance is confirmed by the results of the Wilcoxon signed rank tests ($p$-value$<0.05$). 
This experiment shows that ABX contributes positively to the performance of our EHSA-MTRPP algorithm. 

\subsection{Compared to the route based crossover (RBX) in the literature}\label{subsec:Analyse:rbx}

In this section, we make a comparison between the arc-based crossover and the route-based crossover, which is employed in the reference algorithm MA-MTRPP \cite{lu2019memetic}. 
We create a EHSA-MTRPP variant (EHSA-MTRPP-RBX) by replacing ABX with RBX (line 16 in Algorithm~\ref{Algo:EHSA-MTRPP}).  On each instance of large size, both algorithms are independently run 10 times using the parameters in Section~\ref{subsec:Experi:setup}. The cut-off time is always set to be twice the number of customers. The results are summarized in Table~\ref{tab:rbx}, which uses the same column headings as Table~\ref{tab:nofast}. (Better results are indicated in bold.)
\begin{table}[htbp]
\centering
\renewcommand{\arraystretch}{1.5}
\setlength{\tabcolsep}{0.4mm}
\caption{Results of EHSA-MTRPP-RBX and EHSA-MTRPP on the instances of large size from the benchmark. Each instance is solved 10 times and the cutoff time is set to be twice the number of customers.}{\smallskip}\label{tab:rbx}
\begin{tiny}

\begin{tabular}{llclllclllclllll}
\toprule
\multirow{2}{*}{Problem set}& \multirow{2}{*}{UB}  & & \multicolumn{3}{c}{EHSA-MTRPP-RBX} & & \multicolumn{3}{c}{EHSA-MTRPP} & & \multirow{2}{*}{p-value} & \multirow{2}{*}{$\delta$} & \multirow{2}{*}{W}& \multirow{2}{*}{M}& \multirow{2}{*}{F} \\
\cline{4-6}
\cline{8-10}

 & & &\multicolumn{1}{l}{Best}& \multicolumn{1}{l}{Average} & \multicolumn{1}{l}{Tavg} & & \multicolumn{1}{l}{Best} &\multicolumn{1}{l}{Average} & \multicolumn{1}{l}{Tavg} & & &\\  
\midrule

Ins\_Avci &  &  & &  & & & & & & & & &	&  &  \\ 
\hline 
Size=200, m=2     & 907250.15  &  &			893420.60 & 893254.29 &   156.76& &			\textbf{893513.85} & \textbf{893374.88} &   263.23& &			2.35$\times 10^{-4}$ & 0.010437\% &			19    & 0     & 1     \\ 
Size=200, m=3     & 917633.35  &  &			907886.65 & 907784.43 &   154.88& &			\textbf{907950.35} & \textbf{907841.50} &   258.94& &			1.28$\times 10^{-3}$ & 0.007016\% &			18    & 1     & 1     \\ 
Size=500, m=10    & 1523086.90  &  &			1436503.90 & 1435606.38 &   379.31& &			\textbf{1437256.40} & \textbf{1436265.76} &   898.97& &			5.06$\times 10^{-3}$ & 0.052384\% &			10    & 0     & 0     \\ 
Size=500, m=20    & 766209.10  &  &			694289.30 & 693986.20 &   406.29& &			\textbf{694406.60} & \textbf{694114.40} &   897.70& &			5.06$\times 10^{-3}$ & 0.016895\% &			10    & 0     & 0     \\ 
Size=750, m=100   & 4150788.40  &  &			4000559.80 & 4000520.52 &   641.77& &			\textbf{4000585.60} & \textbf{4000541.60} &  1352.60& &			2.25$\times 10^{-1}$ & 0.000645\% &			4     & 0     & 1     \\ 
Size=1000, m=50    & 5402658.80  &  &			5191587.40 & 5191416.32 &   751.44& &			\textbf{5191726.40} & \textbf{5191527.76} &  1757.60& &			4.31$\times 10^{-2}$ & 0.002677\% &			5     & 0     & 0     \\ 
\midrule
Ins\_Lu &  &  & &  & & & & & & & & &	&  &  \\ 
\hline
 
Size=200, m=2     & 489139.40 &  &			472466.60 & 472286.13 &   300.66& &			\textbf{472499.25} & \textbf{472354.94} &   254.16& &			3.76$\times 10^{-3}$ & 0.006911\% &			17    & 1     & 2     \\ 
Size=200, m=3     & 333141.60 &  &			321230.30 & 321127.52 &   297.78& &			\textbf{321278.75} & \textbf{321175.57} &   245.66& &			3.40$\times 10^{-4}$ & 0.015083\% &			18    & 1     & 1     \\ 
Size=200, m=4     & 246431.05 &  &			236781.60 & 236703.33 &   302.65& &			\textbf{236805.20} & \textbf{236720.93} &   229.22& &			7.37$\times 10^{-4}$ & 0.009967\% &			16    & 2     & 2     \\  

\bottomrule
\end{tabular}
\end{tiny}
\end{table}

The results show that except several cases, EHSA-MTRPP performs better than EHSA-MTRPP-RBX on all sets of instances in terms of the best found results (column `Best') and the average found results (column `Average'), and the Wilcoxon signed rank tests ($p$-value$<0.05$) indicate that there are significant differences for 8 sets of results (except for the instances of `Size=750, m=100'). This experiment confirms that the ABX crossover is more appropriate than RBX for the MTRPP.

\subsection{Rationale behind the arc-based crossover} 
\label{subsec:Ana:rational}


We experimentally investigate the rationale behind ABX by analyzing the structural similarities between high-quality solutions. For two given solutions $\varphi_1$ and $\varphi_2$ with their corresponding arc sets $A_1$ and $A_2$, their similarity is defined by $Sim(\varphi_1, \varphi_2)=\frac{|A_1\cap A_2|}{|A_1\cup A_2|}$. Generally, the larger the similarity between two solutions, the more arcs they share. 

\begin{figure}[htbp]
	\centering
	\includegraphics[scale=0.3]{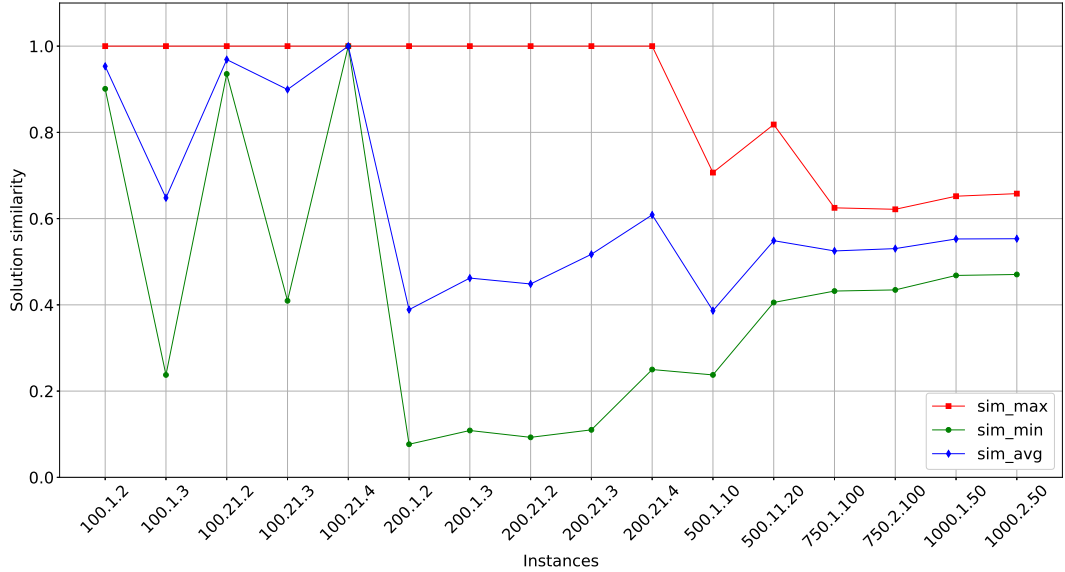}
	\caption{The similarity between high-quality solutions for 16 instances. Each instance is solved 100 times independently with a cut-off time per run set to twice the number of customers.}
	\label{fig:curve2}
\end{figure}

We run the EHSA-MTRPP algorithm 100 times on each of the selected 16 instances while recording the best found solution of each run, and the cut-off time per run is always set to twice the number of customers. For each instance, we calculate the maximum similarity (denoted $sim\_max$) between any two solutions by $sim\_max=\max_{1\leq i< j\leq 100} Sim(\varphi_i, \varphi_j)$, the minimum similarity (denoted $sim\_min$) between  any two solutions by $sim\_min=\min_{1\leq i< j\leq 100} Sim (\varphi_i, \varphi_j)$ and the average similarity (denoted $sim\_avg$) between any two solutions by $sim\_avg=\frac{1}{4950}\cdot\sum_{1\leq i< j\leq 100} Sim(\varphi_i, \varphi_j)$. Figure~\ref{fig:curve2} shows the results of the solution similarities for different instances.

From Figure~\ref{fig:curve2}, one observes that there is a high similarity between high-quality solutions. In particular, the maximum similarity of the 100 high-quality solutions is more than 0.6 and the average similarity is over 0.4 between the solutions for each instance. In other words, a large number of arcs frequently appear in high-quality solutions, which provides a solid foundation for the design of the arc-based crossover in this work.
One notices that the maximum similarities for the last six largest instances ($n\geq 500$) are not as high as the other instances ($n\leq 200$). This may be that the results for these difficult instances ($n\geq 500$) are not good enough. 

\section{Conclusions}\label{sec:Conclu}

An effective hybrid search algorithm for the multiple traveling repairman problem with profit was proposed under the framework of memetic algorithm. 
The proposed algorithm distinguishes itself from the existing algorithms by two key features, i.e., its fast neighborhood evaluation techniques designed to accelerate neighborhood examinations and the dedicated arc-based crossover able to generate diversified and meaningful offspring solutions.



The assessment on the 470 benchmark instances in the literature showed that the proposed algorithm competes favorably with the existing algorithms by updating the best records (new lower bounds) for 137 instances (29\%) and matching the best-known results for 330 instances (70\%) within a reasonable time. Additional experiments demonstrated that the fast evaluation technique and the arc-based crossover play positives roles to the performance of the algorithm. We analyzed both formally and experimentally the reduced complexities of neighborhood examinations and provided experimental evidences (high similarity between the high-quality solutions) to support the design of the arc-based crossover. The source code of our algorithm will be made available upon the publication of this work. It can be used to solve practical applications and adapted to related problems. In the future, we would like to develop efficient algorithms based on the arc-based crossover on some other related problems like team orienteering problem.




\section*{Acknowledgement}

This work was partially supported by the Shenzhen Science and Technology Innovation Commission (grant no. JCYJ20180508162601910), the National Key R\&D Program of China (grant no. 2020YFB1313300), 
and the Funding from the Shenzhen Institute of Artificial Intelligence and Robotics for Society (grant no. AC01202101036 and AC01202101009).

\appendix

\section{Proof of the complexity of neighborhood exploration} \label{app1:proof}

\textbf{Proof of Proposition 1} (Section \ref{subsec:Algo:VNS}). Pei et al. \cite{pei_solving_2020} has proved that for the related TRPP, evaluating one neighboring solution of $Insert$, $2\mbox{-}opt$, $Or\mbox{-}opt$, $Add$ and $Drop$ can be finished in $O(1)$ by using some specific data structures. It is easy to find that these conclusions are equally applicable in the MTRPP because the operations involved in $N_1\mbox{-}N_4$ as well as $N_{Add}$ and $N_{Drop}$  are confined inside one route, which is the same situation as in the TRPP.

For the $Swap$ operator, the proof is given as follows.
Let  $\varphi$ be a solution composed of $K$ routes $\{X_1, X_2, ..., X_{K}\}$, where $X_k=(x_0^k, ..., x_{i-1}^k, x_{i}^k, x_{i+1}^k, ..., x_{j-1}^k,$ $x_j^k,  x_{j+1}^k, ...,  x_{m_k}^k)$ is one route with $m_k$ selected customers.
Swapping $x_i^k$ and $x_j^k$ ($0<i<j\leq m_k$) leads to a neighboring solution $\varphi^\prime$ whose $k$-th route is $X_{k}^\prime=(x_0^k, ..., x_{i-1}^k, x_{j}^k, x_{i+1}^k, ..., x_{j-1}^k, x_i^k,  x_{j+1}^k, ..., x_{m_k}^k )$.
Using Equation (\ref{eq:fast}), the move gain $\Delta_f = f(\varphi^\prime)-f(\varphi)$ can be reached by
\begin{itemize}
\item[1)] If $x_i^k$ and $x_j^k$ are not adjacent, then
\begin{equation*}
\begin{aligned}
\Delta_f
& = (m_k-i+1)\cdot (d_{x_{i-1}^k,x_{i}^k}-d_{x_{i-1}^k,x_{j}^k})+(m_k-i)\cdot (d_{x_{i}^k,x_{i+1}^k}-d_{x_{j}^k,x_{i+1}^k})\\
& + (m_k-j+1)\cdot (d_{x_{j-1}^k,x_{j}^k}-d_{x_{j-1}^k,x_{i}^k})+(m_k-j)\cdot (d_{x_{j}^k,x_{j+1}^k}-d_{x_{i}^k,x_{j+1}^k})
\end{aligned}
\end{equation*}
\item[2)] If $x_i^k$ and $x_j^k$ are adjacent, then
\begin{equation*}
\begin{aligned}
\Delta_f & = (m_k-i+1)\cdot (d_{x_{i-1}^k,x_{i}^k}-d_{x_{i-1}^k,x_{j}^k})+(m_k-j)\cdot (d_{x_{j}^k,x_{j+1}^k}-d_{x_{i}^k,x_{j+1}^k})
\end{aligned}
\end{equation*}
\end{itemize}
Thus, any neighboring solution in $N_1$ can be evaluated in $O(1)$ and the complexity of examining the $N_1$ neighborhood is bounded by $O(m^2)$.

\textbf{Proof of Proposition 2 }(Section \ref{subsec:Algo:VNS}).
Let  $\varphi$ be a solution composed of $K$ routes $\{X_1, X_2, ..., X_{K}\}$, where $X_k=(x_0^k, x_{1}^k, ...,  x_{m_k}^k)$ is one route with $m_k$ selected customers ($k=1, 2, ..., K$). As the set of selected customers does not change, we only consider the change of the accumulated distance according to Equation~(\ref{eq:fast}).

\begin{itemize}
\item[a)]
For the $Inter\mbox{-}Swap$ neighborhood, we suppose two routes in the solution $\varphi$ $$X_a=(x_0^a, x_{1}^a, ...,  x_{i-1}^a, x_{i}^a, x_{i+1}^a, ..., x_{m_a}^a)$$ 
$$X_b=(x_0^b, x_{1}^b, ...,  x_{j-1}^b, x_{j}^b, x_{j+1}^b, ..., x_{m_b}^b)$$ 
Exchanging $x_i^a$ ($0<i\leq m_a$) and $x_j^b$ ($0<j\leq m_b$) leads to a new solution $\varphi^\prime$ whose $a$-th and $b$-th routes are: 
$$X_a^\prime=(x_0^a, x_{1}^a, ...,  x_{i-1}^a, x_{j}^b, x_{i+1}^a, ..., x_{m_a}^a)$$ 
$$X_b^\prime=(x_0^b, x_{1}^b, ...,  x_{j-1}^b, x_{i}^a, x_{j+1}^b, ..., x_{m_b}^b)$$
By Equation~(\ref{eq:fast}), the move gain $\Delta_f = f(\varphi^\prime)-f(\varphi)$ can be achieved by
\begin{equation*}
\begin{aligned}
\Delta_f
& = (m_a-i+1)\cdot (d_{x_{i-1}^a,x_{i}^a}-d_{x_{i-1}^a,x_{j}^b})+(m_a-i)\cdot (d_{x_{i}^a,x_{i+1}^a}-d_{x_{j}^b,x_{i+1}^a})\\
& + (m_b-j+1)\cdot (d_{x_{j-1}^b,x_{j}^b}-d_{x_{j-1}^b,x_{i}^a})+(m_b-j)\cdot (d_{x_{j}^b,x_{j+1}^b}-d_{x_{i}^a,x_{j+1}^b})
\end{aligned}
\end{equation*}
Therefore, each neighboring solution in $N_4$ can be evaluated in $O(1)$, leading to the complexity of $O(m^2)$ for exploring the $Inter\mbox{-}Swap$ neighborhood.

\item[b)] 
For the $Inter\mbox{-}Insert$ neighborhood, we are given two routes for the solution $\varphi$.
$$X_a=(x_0^a, x_{1}^a, ...,  x_{i-1}^a, x_{i}^a, x_{i+1}^a, ..., x_{m_a}^a)$$ 
$$X_b=(x_0^b, x_{1}^b, ...,  x_{j}^b, x_{j+1}^b, ..., x_{m_b}^b)$$ 
Inserting $x_{i}^a$ ($0<i\leq m_a$) into the position between $x_{j}^b$ and $x_{j+1}^b$ ($0 \leq b \leq m_b$) produces a neighboring solution $\varphi^\prime$, whose two corresponding routes are: 
$$X_a^\prime=(x_0^a, x_{1}^a, ...,  x_{i-1}^a, x_{i+1}^a, ..., x_{m_a}^a)$$ 
$$X_b^\prime=(x_0^b, x_{1}^b, ...,  x_{j}^b, x_{i}^a, x_{j+1}^b, ..., x_{m_b}^b)$$
The move gain $\Delta_f$ can be obtained by
\begin{equation*}
\begin{aligned}
\Delta_f
& = 
Vsd_a(i-1)+(m_a+1-i)\cdot d_{x_{i-1}^a, x_i^a}+(m_a-i)\cdot(d_{x_{i}^a, x_{i+1}^a}-d_{x_{i-1}^a, x_{i+1}^a})\\
& -Vsd_b(j)-(m_b+1-j)\cdot d_{x_j^b, x_i^a} - (m_b-j)\cdot(d_{x_i^a, x_{j+1}^a}- d_{x_j^b, x_{j+1}^b})
\end{aligned}
\end{equation*} 
where $Vsd_a(i)$ and $Vsd_b(i)$ are two auxiliary arrays used to accelerate the evaluation procedure.
For the $k$-th route in the solution, the auxiliary array is defined as follows.
\begin{equation}\label{eq:data}
	Vsd_k(i)=\sum_{t=1}^{i} d_{x_{t-1}^k, x_{t}^k}.
\end{equation}
The auxiliary arrays in Equation~(\ref{eq:data}) are pre-calculated and updated for each iteration (the complexity of updating these auxiliary arrays is $O(n)$). Therefore, each neighboring solution can be assessed in $O(1)$ while the complete $Inter\mbox{-}Insert$ neighborhood can be examined in $O(m^2)$.  

\item[c)] For the $Inter\mbox{-}2\mbox{-}opt$ neighborhood, we suppose two routes $X_a$ and $X_b$.
$$X_a=(x_0^a, x_{1}^a, ...,  x_{i}^a, x_{i+1}^a, ..., x_{m_a}^a)$$ 
$$X_b=(x_0^b, x_{1}^b, ...,  x_{j}^b, x_{j+1}^b, ..., x_{m_b}^b)$$
Removing two edges ($x_{i}^a, x_{i+1}^a$) and ($x_{j}^b, x_{j+1}^b$) and replacing them with two other edges lead to a new solution $\varphi^\prime$, which has two corresponding routes $X_a^\prime$ and $X_b^\prime$.
$$X_a^\prime=(x_0^a, x_{1}^a, ...,  x_{i}^a, x_{j+1}^b, ..., x_{m_b}^b)$$ 
$$X_b^\prime=(x_0^b, x_{1}^b, ...,  x_{j}^b, x_{i+1}^a, ..., x_{m_a}^a)$$
The move gain $\Delta_f$ can be obtained as follows.
\begin{equation*}
\begin{aligned}
\Delta_f
& = 
(m_a-i)\cdot d_{x_i^a, x_{i+1}^a} -(m_b-j)\cdot d_{x_i^a, x_{j+1}^b} + (m_a-m_b-i+j) \cdot Vsd_a(i)\\
& +(m_b-j)\cdot d_{x_j^b, x_{j+1}^b} - (m_a-i)\cdot d_{x_j^b,  x_{i+1}^a} + (-m_a+ m_b+i-j) \cdot Vsd_b(j)
\end{aligned}
\end{equation*}
The complexity of evaluating one neighboring solution is thus $O(1)$ and exploring the complete $Inter\mbox{-}2\mbox{-}opt$ neighborhood requires $O(m^2)$.
\end{itemize}

\section{Detailed performance evaluation over the 470 benchmark instances between EHSA-MTRPP and the reference algorithms.}
\label{sec:Appendix}

This appendix presents the detailed results of the proposed MA-MTRPP algorithm compared to the reference algorithms (ALNS-MTRPP \cite{avci_adaptive_2019} and MA-MTRPP \cite{lu2019memetic}) over the 470 benchmark instances. 
Table~\ref{tab:detailavci} summarizes detailed comparison between ALNS-MTRPP and EHSA-MTRPP for the 230 Ins\_Avci instances. The first three columns `Instance'\footnote{The name `Size\_index\_k' for each instance indicates respectively the number of customers, the instance index and the number of servers.}, `UB' and `Bestsofar' describe respectively the name for each instance, the upper bound and the best found result reported in \cite{avci_adaptive_2019}. Then columns `Best', `Average' and `Tavg' present the best found result, the average found result and the average running time to obtain the best found result of ALNS-MTRPP for each instance. The following three columns list the same information for our EHSA-MTRPP algorithm. The last column `imp' gives the improvement of EHSA-MTRPP over ALNS-MTRPP in terms of the best found solution. (Better results are indicated in bold.)
Using similar column headings to Table~\ref{tab:detailavci} (excluding column `Bestsofar'), Table~\ref{tab:detaillu} gives the detailed comparison between MA-MTRPP and EHSA-MTRPP for each of the 240 Ins\_Lu instances. (Better results are also marked in bold.)

\renewcommand{\arraystretch}{1.3}
\setlength\tabcolsep{1.5pt}

\begin{center}
	\begin{tiny}
\begin{longtable}[H]{@{\extracolsep{\fill}}lllclllclllcl}

		\caption{Detailed results obtained by the reference algorithm ALNS-MTRPP \cite{avci_adaptive_2019} and our algorithm EHSA-MTRPP over the 230 Ins\_Avci instances, where each instance was solved 5 times according to \cite{avci_adaptive_2019}. The optimal solutions of the instances for $Size=10$ are known but their timing information for ALNS-MTRPP is not available. Improved results are indicated in bold.}\label{tab:detailavci}\\[-2mm]
		\hline \noalign{\smallskip}
		
\multirow{2}{*}{Instance}& \multirow{2}{*}{UB}& \multirow{2}{*}{Bestsofar}  & & \multicolumn{3}{c}{ALNS-MTRPP} & & \multicolumn{3}{c}{EHSA-MTRPP} & &\multirow{2}{*}{imp}  \\
\cline{5-7}
\cline{9-11}

 & & &  &\multicolumn{1}{l}{Best}& \multicolumn{1}{l}{Average} & \multicolumn{1}{l}{Tavg} & & \multicolumn{1}{l}{Best} &\multicolumn{1}{l}{Average} & \multicolumn{1}{l}{Tavg} & & \\  
 
		\hline \noalign{\smallskip}
		
		\endfirsthead
		
		\multicolumn{10}{l}{{\footnotesize\tablename\ \thetable{} -- Continued from previous page}}\\[-3mm]
		\multicolumn{12}{c}{}\\
		\hline \noalign{\smallskip}
\multirow{2}{*}{Instance}& \multirow{2}{*}{UB}& \multirow{2}{*}{Bestsofar}  & & \multicolumn{3}{c}{ALNS-MTRPP} & & \multicolumn{3}{c}{EHSA-MTRPP} & &\multirow{2}{*}{imp}  \\
\cline{5-7}
\cline{9-11}

 & & &  &\multicolumn{1}{l}{Best}& \multicolumn{1}{l}{Average} & \multicolumn{1}{l}{Tavg} & & \multicolumn{1}{l}{Best} &\multicolumn{1}{l}{Average} & \multicolumn{1}{l}{Tavg} & &\\  
		
		\hline \noalign{\smallskip}
		
		\endhead

		\hline \multicolumn{10}{l}{Continued on next page ...}\\ \hline  
		\endfoot
		\hline \noalign{\smallskip}
		\hline \noalign{\smallskip}
		
		\endlastfoot		
10\_1\_2 &     2975 &     2975 &  &			    2975 &  2975.00 &        *& &			    2975 &  2975.00 &     0.03& &			       0 \\ 
10\_2\_2 &     2319 &     2319 &  &			    2319 &  2319.00 &        *& &			    2319 &  2319.00 &     0.02& &			       0 \\ 
10\_3\_2 &     1997 &     1997 &  &			    1997 &  1997.00 &        *& &			    1997 &  1997.00 &     0.03& &			       0 \\ 
10\_4\_2 &     2688 &     2688 &  &			    2688 &  2688.00 &        *& &			    2688 &  2688.00 &     0.02& &			       0 \\ 
10\_5\_2 &     2644 &     2644 &  &			    2644 &  2644.00 &        *& &			    2644 &  2644.00 &     0.03& &			       0 \\ 
10\_6\_2 &     2207 &     2207 &  &			    2207 &  2207.00 &        *& &			    2207 &  2207.00 &     0.02& &			       0 \\ 
10\_7\_2 &     1483 &     1483 &  &			    1483 &  1483.00 &        *& &			    1483 &  1483.00 &     0.03& &			       0 \\ 
10\_8\_2 &     2060 &     2060 &  &			    2060 &  2060.00 &        *& &			    2060 &  2060.00 &     0.02& &			       0 \\
\end{longtable}
\end{tiny}
\end{center}


\renewcommand{\arraystretch}{1.3}
\setlength\tabcolsep{1.5pt}

\begin{center}
	\begin{tiny}
	\begin{longtable}[H]{@{\extracolsep{\fill}}llclllclllcl}

		\caption{Detailed results obtained by the reference algorithm MA-MTRPP \cite{lu2019memetic} and our algorithm MA-MTRPP over the 240 Ins\_Lu instances, where each instance was solved 10 times according to \cite{lu2019memetic}. The optimal solutions of the instances for $Size=20$ are known. Improved results are indicated in bold.}\label{tab:detaillu}\\[-2mm]
		\hline \noalign{\smallskip}
		
\multirow{2}{*}{Instance}& \multirow{2}{*}{UB}  & & \multicolumn{3}{c}{MA-MTRPP} & & \multicolumn{3}{c}{EHSA-MTRPP} & &\multirow{2}{*}{imp}  \\
\cline{4-6}
\cline{8-10}

 & &  &\multicolumn{1}{l}{Best}& \multicolumn{1}{l}{Average} & \multicolumn{1}{l}{Tavg} & & \multicolumn{1}{l}{Best} &\multicolumn{1}{l}{Average} & \multicolumn{1}{l}{Tavg} & & \\  
 
		\hline \noalign{\smallskip}
		
		\endfirsthead
		
		\multicolumn{10}{l}{{\footnotesize\tablename\ \thetable{} -- Continued from previous page}}\\[-3mm]
		\multicolumn{12}{c}{}\\
		\hline \noalign{\smallskip}
\multirow{2}{*}{Instance}& \multirow{2}{*}{UB}  & & \multicolumn{3}{c}{MA-MTRPP} & & \multicolumn{3}{c}{EHSA-MTRPP} & &\multirow{2}{*}{imp}  \\
\cline{4-6}
\cline{8-10}

 & &  &\multicolumn{1}{l}{Best}& \multicolumn{1}{l}{Average} & \multicolumn{1}{l}{Tavg} & & \multicolumn{1}{l}{Best} &\multicolumn{1}{l}{Average} & \multicolumn{1}{l}{Tavg} & & \\  
		
		\hline \noalign{\smallskip}
		
		\endhead

		\hline \multicolumn{10}{l}{Continued on next page ...}\\ \hline  
		\endfoot
		\hline \noalign{\smallskip}
		\hline \noalign{\smallskip}
		
		\endlastfoot		
20\_21\_2 &     4041 &  &				    4041 &  4041.00 &     1.30& &				    4041 &  4041.00 &     0.06& &				       0 \\ 
20\_22\_2 &     4309 &  &				    4309 &  4309.00 &     1.30& &				    4309 &  4309.00 &     0.06& &				       0 \\ 
20\_23\_2 &     4800 &  &				    4800 &  4800.00 &     1.50& &				    4800 &  4800.00 &     0.05& &				       0 \\ 
20\_24\_2 &     2442 &  &				    2442 &  2442.00 &     1.40& &				    2442 &  2442.00 &     0.05& &				       0 \\ 
20\_25\_2 &     3872 &  &				    3872 &  3872.00 &     1.20& &				    3872 &  3872.00 &     0.04& &				       0 \\ 
20\_26\_2 &     4067 &  &				    4067 &  4067.00 &     1.40& &				    4067 &  4067.00 &     0.05& &				       0 \\ 
20\_27\_2 &     3410 &  &				    3410 &  3410.00 &     1.30& &				    3410 &  3410.00 &     0.05& &				       0 \\ 
20\_28\_2 &     3721 &  &				    3721 &  3721.00 &     1.30& &				    3721 &  3721.00 &     0.05& &				       0 \\
\end{longtable}
\end{tiny}
\end{center}

\end{document}